\documentclass[conference]{IEEEtran}
\IEEEoverridecommandlockouts
\usepackage{cite}
\usepackage{amsmath,amssymb,amsfonts}
\usepackage{algorithmic}
\usepackage{graphicx}
\usepackage{textcomp}
\usepackage{xcolor}
\usepackage{hyperref}
\usepackage{url}
\usepackage{amsmath,amssymb,amsfonts}
\usepackage{graphicx}
\usepackage{textcomp}
\usepackage{xcolor}

\def\BibTeX{{\rm B\kern-.05em{\sc i\kern-.025em b}\kern-.08em
    T\kern-.1667em\lower.7ex\hbox{E}\kern-.125emX}}
    
\usepackage{tikz}
\usepackage{amsmath}
\usepackage{tikz}
\usepackage{amsmath}

\usepackage{filecontents}
\usepackage{microtype}
\usepackage{graphicx}
\usepackage{subfigure}
\usepackage{booktabs} % for professional tables
\usepackage{hyperref}
\usepackage{algorithm}
\usepackage{algorithmic}

\usepackage{wrapfig}
\usepackage{amsmath}
\DeclareMathOperator*{\agg}{AGG}

\usepackage{amssymb}
\usepackage{mathtools}
\usepackage{amsthm}

\usepackage{filecontents}
\usepackage{microtype}
\usepackage{graphicx}
\usepackage{subfigure}
\usepackage{booktabs} % for professional tables
\usepackage{hyperref}
\usepackage{algorithm}
\usepackage{algorithmic}

\usepackage{wrapfig}
\usepackage{amsmath}
\usepackage{amssymb}
\usepackage{mathtools}
\usepackage{amsthm}
\usepackage[capitalize,noabbrev]{cleveref}

\usepackage{soul} 
\theoremstyle{plain}
\newtheorem{theorem}{Theorem}

\newtheorem{lemma}[theorem]{Lemma}

\theoremstyle{definition}
\newtheorem{definition}[theorem]{Definition}

\theoremstyle{remark}
\DeclareMathOperator*{\argmax}{arg\,max}

\def\BibTeX{{\rm B\kern-.05em{\sc i\kern-.025em b}\kern-.08em
    T\kern-.1667em\lower.7ex\hbox{E}\kern-.125emX}}
\begin{document}
\IEEEoverridecommandlockouts
\IEEEpubid{\makebox[\columnwidth]{978-1-6654-8045-1/22/\$31.00 \copyright2022 IEEE \hfill} \hspace{\columnsep}\makebox[\columnwidth]{}}
\title{Heterogeneous Randomized Response for Differential Privacy in Graph Neural Networks
\thanks{$^*$ \textit{Corresponding author}}
}

\author{
% Anonymous author(s)
\IEEEauthorblockN{Khang Tran, Phung Lai, NhatHai Phan$^*$}
\IEEEauthorblockA{
% \textit{Information System Department} \\
\textit{New Jersey Institute of Technology, USA} \\
\{kt36, tl353, phan\}@njit.edu}
\and
\IEEEauthorblockN{Issa Khalil}
\IEEEauthorblockA{\textit{Qatar Computing Research Institute, Qatar} \\
ikhalil@hbku.edu.qa}
\and
\IEEEauthorblockN{Yao Ma, Abdallah Khreishah}
\IEEEauthorblockA{
% \textit{Computer Science Department} \\
\textit{New Jersey Institute of Technology, USA} \\
\{yao.ma, abdallah\}@njit.edu
}
\and
\IEEEauthorblockN{My T. Thai}
\IEEEauthorblockA{
\textit{University of Florida, USA}\\
mythai@cise.ufl.edu}
\and
\IEEEauthorblockN{Xintao Wu}
\IEEEauthorblockA{
% \textit{Computer Science and Computer Engineering Department} \\
\textit{University of Arkansas, USA} \\
xintaowu@uark.edu}
}

\maketitle

\begin{abstract}
Graph neural networks (GNNs) are susceptible to privacy inference attacks (\textsc{PIAs}) given their ability to learn joint representation from features and edges among nodes in graph data.
To prevent privacy leakages in GNNs, we propose a novel heterogeneous randomized response (\textsc{HeteroRR}) mechanism to protect nodes' features and edges against \textsc{PIAs} under differential privacy (DP) guarantees, without an undue cost of data and model utility in training GNNs.
Our idea is to balance the importance and sensitivity of nodes' features and edges in redistributing the privacy budgets since some features and edges are more sensitive or important to the model utility than others. 
As a result, we derive significantly better randomization probabilities and tighter error bounds at both levels of nodes' features and edges departing from existing approaches, thus enabling us to maintain high data utility for training GNNs.
An extensive theoretical and empirical analysis using benchmark datasets shows that \textsc{HeteroRR} significantly outperforms various baselines in terms of model utility under rigorous privacy protection for both nodes' features and edges. That enables us to defend PIAs in DP-preserving GNNs effectively.
\end{abstract}


\section{Introduction}
Graph Neural Networks (GNNs) have been well-known for their ability to learn from graph data, simultaneously leveraging the nodes' features and the graph structure \cite{gnnSurvey}. However, GNNs are vulnerable to PIAs since the nodes' features and the edges often contain sensitive information of the participants, which can be inferred by the adversaries when GNNs are deployed \cite{privateLeakageQuantifying}. Attacks such as membership inference \cite{membershipInferenceAttack}, and structure inference \cite{linkTeller} underline privacy risks in GNNs. 
Hence, to promote the broader adoption of GNNs, it is essential to protect graph data privacy in training GNNs while maintaining high model utility.

% \renewcommand{\thefootnote}{\fnsymbol{footnote}}
% \footnotetext{$^*$\textit{Corresponding author}}

% To protect the privacy of graph data, a privacy-preserving method must protect both sensitive nodes' features and edges in training GNNs while retaining high data and model utility.

Among privacy preserving techniques, differential privacy (DP), a rigorous formulation of privacy in probabilistic terms without computational overhead, is one of the golden standards. DP has been applied to protect either edge privacy \cite{edgeprivacy} or nodes' feature privacy \cite{locallyprivateGNN} given graph data. To protect both the graph structure and node features, a straightforward approach to achieve DP protection at both nodes' feature-level and graph structure-level in training GNNs is applying both node-feature and graph-structure DP-preserving mechanisms independently. However, that treatment can significantly degrade the graph data utility resulting in poor model performance, especially in the application of GNNs. This is a challenging and open problem since 
a minor privacy-preserving perturbation to either features of a single node or a local graph structure will negatively affect its neighbors. In addition, the impact is propagated through the entire graph. There are two main reasons for this problem: first, the correlation between the nodes is very high, therefore, quantifying the privacy risk through the aggregation of GNNs is intractable and adding a little noise to one node can impact all of its neighbors; second, most of graphs in practice are sparse and adding noise to the structure of the graph can easily destroy the sparsity of it, resulting in low graph structure utility.
% The root cause of this problem is the strong correlation between the nodes and the sparsity of the graph structure.
% \st{Addressing these problems is challenging and it demands a deep understanding regarding the correlation among graph data to optimize the trade-off of privacy and data utility.} 

\textbf{Key Contributions.} To address this problem, we develop a new heterogeneous randomized response (\textsc{HeteroRR}) mechanism to preserve nodes' features and edges privacy for graph data in GNNs application. Our methods are based on randomize response (RR) 
% \footnote{A mechanism allows correct probabilistic responses to a sensitive issue while maintaining confidentiality by plausible deniability of the responses. A basic example of RR is a survey to find out the average number of people who smoke in which each participant is asked, "Do you smoke?". Each participant can flip a coin. If "head," then answer with the truth. If "tail," then the participants flip the coin a gain and answer "Yes" if "head" or "No" if "tail."} 
\cite{ldp4, ldp5} which is an advanced and effective method for privacy-preserving. \textsc{HeteroRR} leverages the heterogeneity in graph data to optimize the magnitude of privacy preserving-noise injected into nodes' features and the graph structure, such that \textbf{\textit{less sensitive}} and \textbf{\textit{more important}} features (to the model outcome) and edges receive \textbf{\textit{lower probabilities to be randomized (less noisy)}}, and vice versa. This property of \textsc{HeteroRR} enables us to achieve significantly better utility compared with homogeneous randomization probabilities in existing mechanisms under the same privacy guarantee.

Furthermore, \textsc{HeteroRR} is applied in the pre-processing step to create a privacy-preserving graph that can be stored and reused as a replacement for the original graph. Due to the post-processing of DP, every analysis on the privacy-preserving graph satisfies the DP guarantee for the original graph, which makes \textsc{HeteroRR} a permanent RR. Therefore, it provides longitudinal DP protections without accumulation of privacy risks over the time.

An extensive theoretical and empirical analysis conducted on benchmark datasets employing GNNs as a motivating application shows that \textsc{HeteroRR} significantly outperforms baseline approaches in terms of data and model utility under the same privacy protection. Importantly, \textsc{HeteroRR} are resilient against PIAs by reducing the attack success rate to a random guess level without affecting the GNNs' model utility. Our implementation and supplemental documents can be found here: \textit{\url{https://github.com/khangtran2020/DPGNN.git}}

% \textbf{Outline.} The paper is organized as follows. Section 2 reviews graph learning, privacy threat models, and DP-preserving mechanisms in graphs. We present \textsc{HeteroRR} for nodes' feature-level and edge-level in Section 3 and prove its privacy properties in Section 4. We present our experimental results in benchmark datasets to provide insight on why \textsc{HeteroRR} satisfies the rigorous LDP guarantee and achieves good data utility in Section 5. Section 6 concludes the paper and discusses future work. The codes of our implementation and our supplemental documents can be found here: \textit{\url{https://github.com/khangtran2020/DPGNN.git}}

\section{Background}
This section provides an overview of GNNs, privacy threat models, and existing defenses.

% \paragraph{Randomized Response.} \hai{Provide an overview of RR.}

\paragraph{Graph Learning Setting} A service provider possesses a private graph $G(\mathcal{V}, \mathcal{E})$ constructed from its users' data, where $\mathcal{V}$ is the set of nodes, $\mathcal{E}$ is the set of edges. Each node $v \in \mathcal{V}$ has its raw (data) input $x$ and a \textit{ground-truth} one hot vector $y \in \{0,1\}^\mathcal{C}$ with $\mathcal{C}$ is the number of output classes. 
Each node (user) $v \in \mathcal{V}$ has a set of \textit{public} (non-sensitive) edges and a set of \textit{private} (sensitive) edges (i.e., $\mathcal{E} = \mathcal{E}_{pub} \cup \mathcal{E}_{pri}$). 
This is a practical setting in many real-world applications.

For instance, in a FLICKR network \cite{snapflickr}, an edge between a pair of images (nodes) can be created by a mutual friend connection of the image owners. These edges can expose private friend connections among the image owners; thus, they are sensitive and need to be protected. Meanwhile, an edge between a pair of images constructed based on either shared galleries or common tags is considered public since the edge does not expose private connections among the image owners. 

In practice, one can use a pretrained model $g(\cdot)$ to extract a $d$-dimension embedding vector $z = g(x)$ as an initial representation of each node $v \in \mathcal{V}$. 
% Let $A \in \{0,1\}^{|\mathcal{V}| \times |\mathcal{V}|}$ be the adjacency matrix of $G$ where $A_{u,v} = 1$ if node $u$ connects with node $v$; otherwise, $A_{u,v} = 0$. 
A $K$-layer GNN  learns the embedding representation for each node $v \in \mathcal{V}$ through a stack of $K$ graph convolutional layers. Each layer $k \in [1,K]$ takes as input the embedding $h^{(k-1)}_v$ for $v \in \mathcal{V}$ from the previous layer, then updates the embedding as follows:

{
\small
\begin{align}
\label{eq:gcnupdate}
    h^{(k)}_{\mathcal{N}(v)} &= \agg\big(h^{(k-1)}_v \cup \{h_u^{(k-1)}, u \in \mathcal{N}(v)\}\big)\Big) \\ 
    h^{(k)}_v &= \sigma\Big(W^{(k)}h^{(k)}_{\mathcal{N}(v)}\Big)
\end{align}
}
% \phung{Why Eq2 is $h^{(k)}_v$ in the LHS? if it is the aggregated one at layer $k$ and differs from $h^{(k)}_v$ in the RHS, we should use different notations for  $h^{(k)}_v$ on LHS and RHS.}\khang{I changed. Please check again}
where $\mathcal{N}(v)$ is the neighborhood of node $v$, $h^{(0)}_v = z_v$, $\agg(\cdot)$ is an aggregation function, $W^{(k)}$ is the trainable parameters of layer $k$, and $\sigma(\cdot)$ is a non-linear activation function. 

In this work, we consider a node classification task, in which each node is classified into one of the output classes. At the inference time, the service provider releases APIs to query the trained model in applications. This is a practical setting of ML-as-a-service (MLaaS) for GNNs \cite{linkTeller}.

\begin{figure}[t]
% % \vskip 0.2in
    \centering
    \includegraphics[width=0.88\columnwidth]{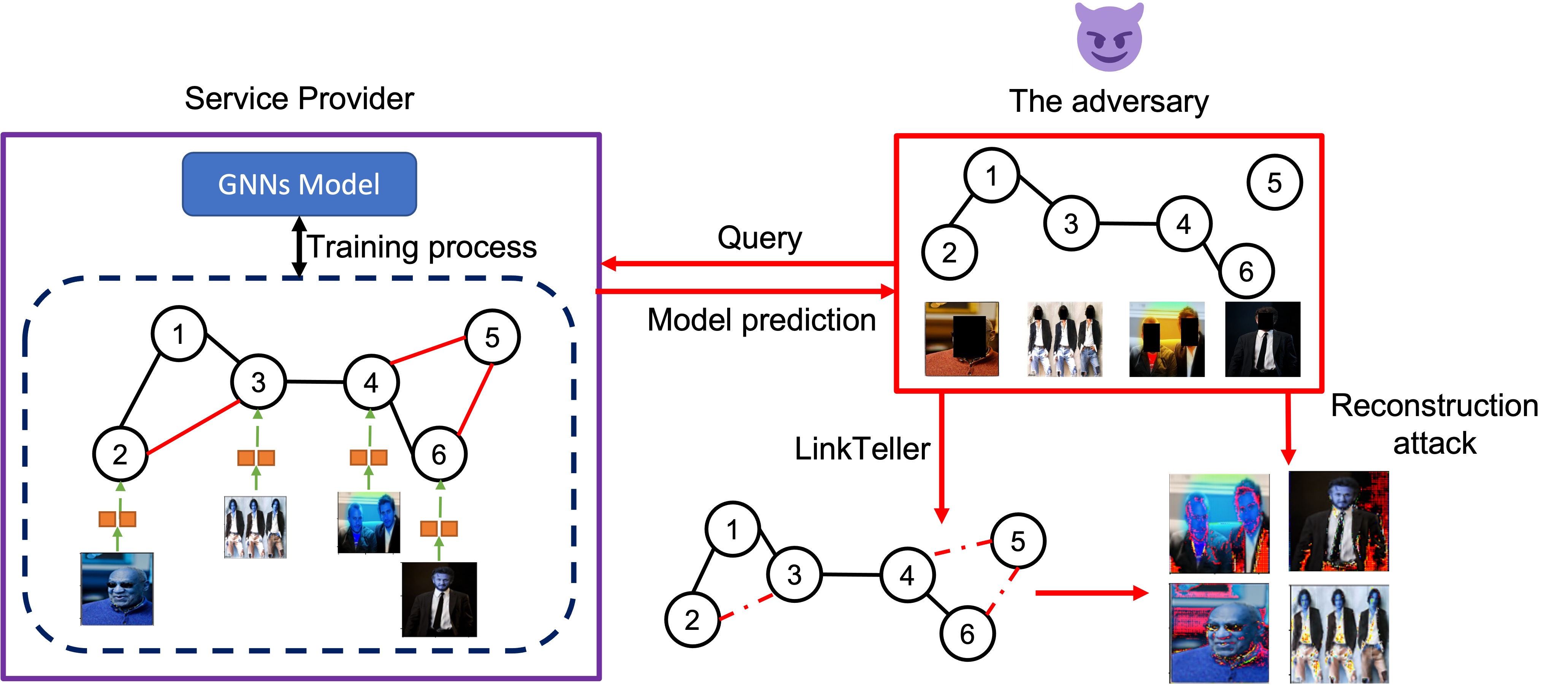} \vspace{-10pt}
    \caption{Privacy threat} 
    \vspace{-20pt}
    \label{fig:adversary}

\end{figure}

\paragraph{Privacy Threat Models} 
Given the graph learning setting, we consider the threat model as in Figure \ref{fig:adversary}. An adversary aims to infer the nodes' raw (data) input and the private connections in the private graph $G$. Firstly, the adversary collects the auxiliary information of the nodes' features and the public edges from the public sources to infer the private connections by conducting the LinkTeller attack \cite{linkTeller}. Secondly, the adversary uses the auxiliary information with the inferred set of edges to perform an inference attack to infer nodes' embedding features \cite{attributeinference}. Finally, the adversary uses the inferred embedding features to reconstruct the raw (data) input. This threat model leads to severe privacy leakages in using sensitive graph data in GNNs.

\paragraph{Differential Privacy (DP) \cite{dwork14}} DP is a privacy-aware computational approach that assure the output of a mechanism is not strongly dependent on a particular data-point. However, in graph data, features of a particular node can be inferred by neighboring nodes' features. To address this problem, one can employ local DP (LDP) to ensure that the privacy-preserving randomization of every each node's features is independently its neighboring nodes. The definition of LDP is as follows:
\begin{definition}[Local DP]
\label{def:LDP}
    A randomized algorithm $\mathcal{M}$ satisfies $\varepsilon$-LDP if and only if  for any pair of inputs $z,z'$ in the input space, and the output space $S\subseteq Range(\mathcal{M})$, it satisfies
    
    {
    \small
    \begin{equation}
      Pr[\mathcal{M}(z)\in S] \le e^\varepsilon Pr[\mathcal{M}(z') \in S]
    \end{equation}
    }
where $\varepsilon$ is the privacy budget. The privacy budget $\varepsilon$ control how the output distribution conditioned by $z$ and $z'$ may differ. A smaller value of $\varepsilon$ ensure a better privacy protection.
\end{definition}

One of effective methods to preserve LDP is applying randomized response (RR) mechanisms \cite{ldp4,ldp5}. Existing RR methods consider a homogeneous scenario where every features in the input space have the same sensitivity and importance which is not utility-optimal since the input space is heterogeneous in most of real-world tasks. Departing from existing approaches, we derive heterogeneous randomization probabilities across features by balancing their sensitivity and importance; thus achieving better graph data privacy-data utility trade-offs in \textsc{HeteroRR}.

\paragraph{DP Preservation in Graph Structure} DP preservation in graph analysis can be generally categorized into node-level DP \cite{nodeDP} and edge-level DP \cite{edgeDP}. Node-level DP and edge-level DP aims to protect the presence of the set of nodes or edges, respectively. In this work, we focus on edge-level DP to protect the privacy of graph structure, defined as follows:

\begin{definition}[Edge-level DP \cite{edgeDP}] A randomized algorithm $\mathcal{M}$ satisfies edge-level $\varepsilon$-DP if for any two neighboring graphs $G(\mathcal{V}, \mathcal{E})$ and $G'(\mathcal{V}, \mathcal{E}')$ differ in one edge while they share the same set of nodes and an output space $\mathcal{O} \subseteq Range(\mathcal{M})$, 

{
\small
\begin{equation}
    P(\mathcal{M}(G) \in \mathcal{O}) \le e^\varepsilon P(\mathcal{M}(G') \in \mathcal{O})
\end{equation}
}
where $\varepsilon$ is the privacy budget. Edge-level DP ensures that the adversary cannot infer the existence of a targeted edge with high confidence.
\end{definition}

Previous edge-level mechanisms consider the whole set of edges is private. That may not be practical in certain real-world scenarios. 
% For instance, in a FLICKR network \cite{snapflickr}, only edges constructed by the connections of their image owners are sensitive. 
Therefore, \textsc{HeteroRR} focuses on protecting the set of private edges and leverages the information from the public edges to optimize the model utility.

% \khang{mention DP for edge level}

%\cite{spectralDP} and \cite{graphmining} preserve the privacy of the graph data in the data analysis scenarios by adding Laplace noise into the queries output of the graph mining task. 
\begin{figure}[t]
% % \vskip 0.2in
    \centering     %%% not \center
    \subfigure[Graph $G(\mathcal{V}, \mathcal{E})$]{\label{fig:a}\includegraphics[width=0.35\columnwidth]{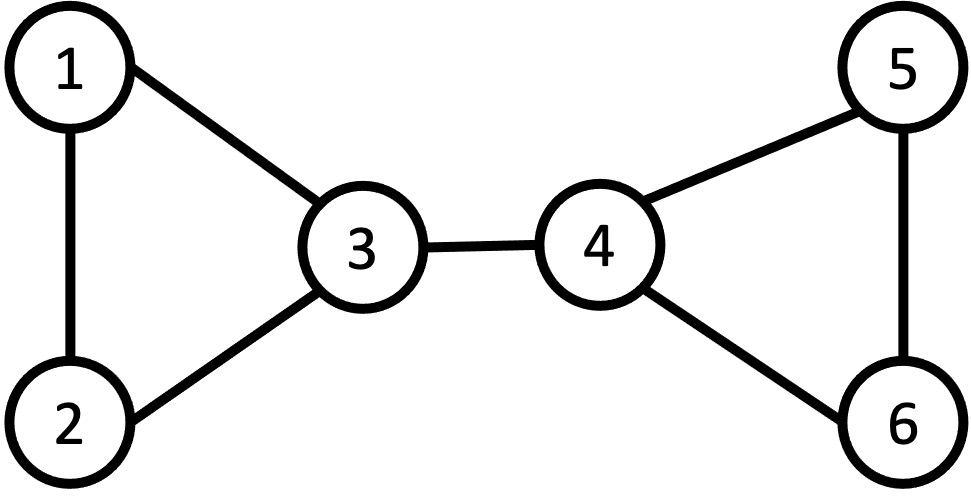}}
    \subfigure[HRG model ($\mathcal{D}$,$\{p_r\}$)]{\label{fig:b}\includegraphics[width=0.55\columnwidth]{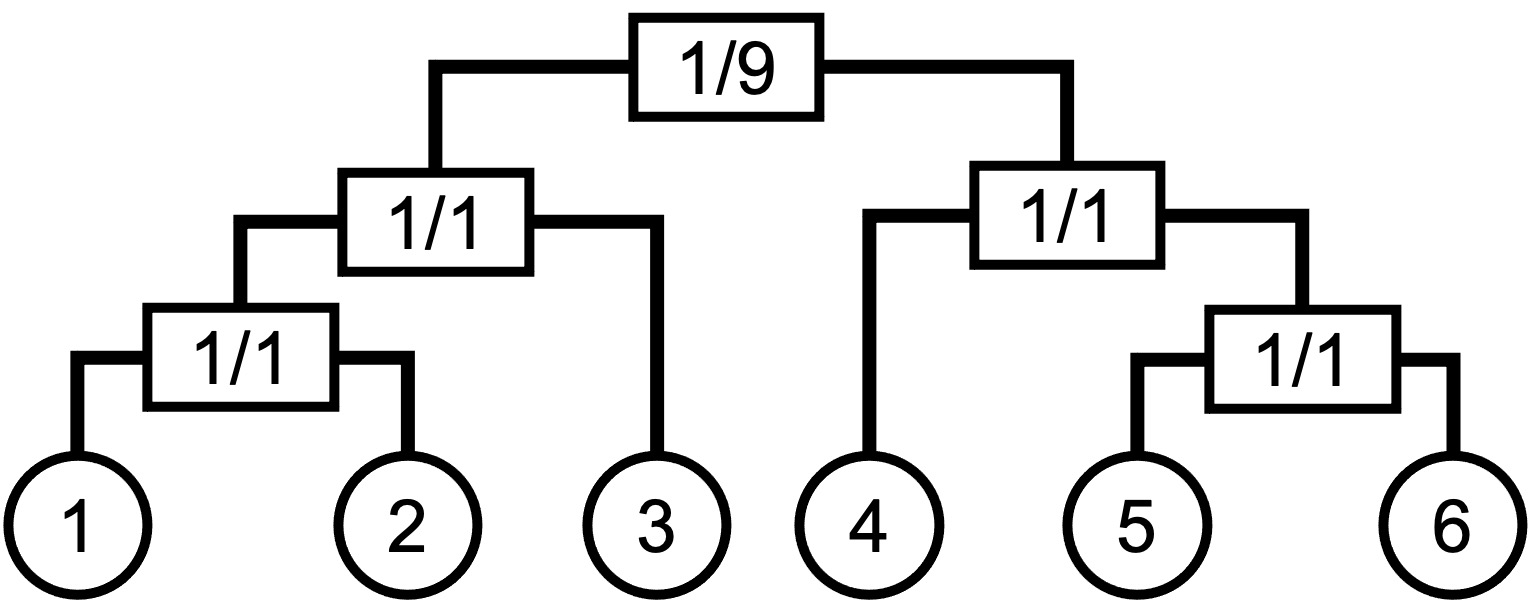}} \vspace{-5pt}
    \caption{An instance of HRG model (b) represents a given graph $G$ (a). Considering the common ancestor of the leaves (1) and (2): since there are one leaf in the left child and one leaf in the right child, the number of possible edges is one. Since there are one edges between node (1)-(2) in $G$, the probability $p_r$ at this internal node $r$ is 1/1.} 
    \vspace{-15pt}
    \label{fig:hrg}
\end{figure}

\paragraph{Hierarchical Random Graph (HRG) \cite{HRG}} To design an edge-level DP-preserving mechanism, one of the state-of-the-art approaches is representing the given graph $G$ as a HRG model \cite{HRG}. HRG is a statistical inference model that represents a hierarchical property of a given graph $G$ by a binary tree dendrogram $D$ as illustrated in Figure \ref{fig:hrg}. In the dendrogram $D$, the number of leaves equal to the number of nodes in $G$. Each internal node $r \in D$ is associated with a probability of having a connection between the left and right child of $r$. Clauset et al. \cite{HRG} proposed using Monte Carlo Markov Chain (MCMC) sampling to find the best HRG model to present a given graph.
Xiao et al. \cite{kddbaseline} proposed to use the exponential and Laplace mechanisms to randomize the sampling process of HRG under DP.
Then the DP-preserving HRG is used to sample a DP-preserving graph $\bar{G}$ which is released to the public. Different from \cite{kddbaseline}, \textsc{HeteroRR} leverages public edges to optimize the structural utility while providing edge-level DP guarantees to protect private edges. We achieve these two objectives in a unified edge-level DP-preserving MCMC sampling algorithm.

\section{\textsc{HeteroRR}: DP preserving in GNNs}

In this section, we formally introduce \textsc{HeteroRR} mechanism to preserve private information of both the nodes' embedding features and the private edges against the aforementioned threat model while maintaining high data utility. 
% Compared with other data domains, such as images, graph data utility is susceptible to privacy-preserving noise, especially in graph-based learning tasks, since either small noise injected into nodes' features or rewiring an edge can significantly degenerate graph data utility.

% We address the problem step by step. 
% Our key idea is to: \textbf{(1)} Derive heterogeneous probabilities for nodes' embedding features such that features with \textit{less sensitivity and more importance to data utility} have \textit{lower probabilities} to be randomized, and vice versa; and \textbf{(2)} Leverage public edges to better retain the graph structure in the randomizing process of the private edges. \hai{The connection with HRG is missing.}

\textbf{Overview of \textsc{HeteroRR}.} \textsc{HeteroRR} consists of two main components \underline{f}eature-\underline{a}ware \underline{r}andomized \underline{r}esponse (\textsc{FeatureRR}) and \underline{e}dge-\underline{a}ware \underline{r}andomized \underline{r}esponse (\textsc{EdgeRR})  which provide nodes' feature-level and edge-level privacy protection respectively. First, \textsc{FeatureRR} randomizes every embedding feature to generate $\varepsilon_f$-LDP-preserving embedding features for every node. Second, \textsc{EdgeRR} represents a given graph with an HRG model under DP protection and then uses the HRG model to sample an $\varepsilon_e$-DP-preserving graph as a replacement for the original graph. Finally, we combine the $\varepsilon_f$-LDP-preserving embedding features and the $\varepsilon_e$-DP-preserving graph structure to create a final and permanent DP preserving graph for training GNNs. 

To optimize the data utility, first, \textsc{FeatureRR} balances the sensitivity and importance of each feature and randomize every feature such that \textit{less sensitive} and \textit{more important} features have \textit{lower probabilities of being randomized} and vice versa. Secondly, in \textsc{EdgeRR}, we propose a novel HRG model to capture the hierarchy of public and private edges in the original graph called and construct the \textsc{pHRG} by iteratively applying a new MCMC sampling process in which (1) we add noise to the MCMC to protect the private edges; and (2) we leverage public edges to preserve the original graph structure. As a result, \textsc{HeteroRR} achieves better data utility under DP protection.

\subsection{Feature-aware Randomized Response (\textsc{FeatureRR})}

% \begin{figure}[t]
% % % \vskip 0.2in
%     \centering
%     \includegraphics[width=0.8\columnwidth]{images/impsens.jpg} \vspace{-5pt}
%     \caption{An overview of feature's importance and sensitivity indication process.}
%     \vspace{-10pt}
%     \label{fig:impsensextract}
% \end{figure}

\begin{figure}[t]
% % \vskip 0.2in
    \centering
    \includegraphics[width=0.8\columnwidth]{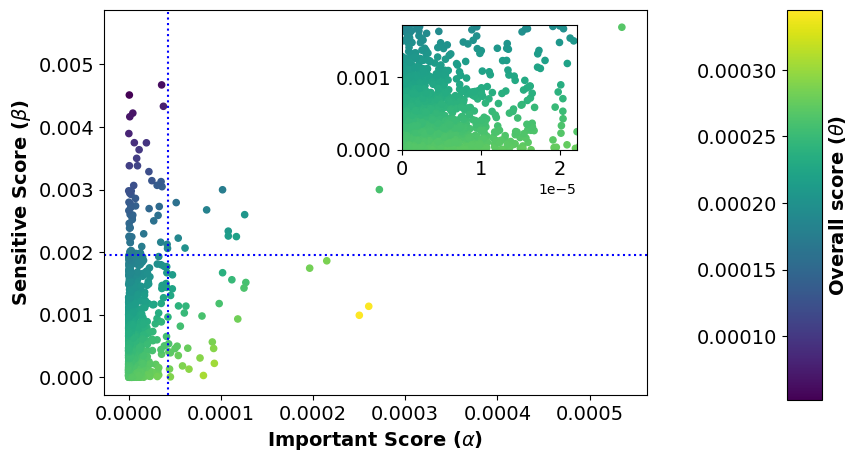} \vspace{-5pt}
    \caption{The overall score $\theta$ with $\gamma = 0.5$}
    \vspace{-15pt}
    \label{fig:noisescale - Hai}
\end{figure}

\paragraph{Sensitivity and Importance} Let us present our method to determine the sensitivity and importance of the embedding features. 
% Figure \ref{fig:impsensextract} (in our supplemental documents) illustrates an overview of our method. 
In the input $x$, some input features are more sensitive or more important to the model outcome than others. 
% For instance, income or disease in healthcare applications or faces in images as considered in this work. 
This triggers a simple question: \textit{``How could we quantify the sensitivity and importance of an embedding feature given an input feature?''}
% \khang{Therefore, in the embedding features $z$, there are more sensitive or important features than others. We define the sensitivity of a feature is how much it changes when the sensitive information is removed from the raw (data) input and the importance of a feature is how much it affects the model's decision.}
To answer this question, first, we quantify the sensitivity of an embedding feature as the maximal magnitude it can be changed by completely removing the sensitive input features 
% in the worse case; that is: When the sensitive input features are completely removed 
from the input $x$. Therefore, we mask all of the sensitive input features out in a masked version $\hat{x}$ and extract its embedding features $\hat{z} = g(\hat{x})$. Then, we quantify the sensitivity $\beta_i$ of an embedding feature $i$ as follows:
{
\small
\begin{align}
    \forall i \in [d]: \beta_i = \frac{|z_i - \hat{z}_i|}{\|z - \hat{z}\|_1} \label{eq:sensscore}
\end{align}
}
% Regarding the importance of an embedding feature, it can be quantified by how much the feature can change the model's decision.

% \rvthree{ shouldn’t sensitivity be defined in the worst-case, for the sake of ensuring DP?}

Regarding the importance of the embedding feature $i$, denoted $\alpha_i$, we employ the SHAP metric \cite{shapscore}, one of the most well-applied model explainers, to quantify the influence of feature $i$ on a model's decision. The importance score is quantified as follows:
{
\small
\begin{align}
    \forall i \in [d]: \alpha_i = \frac{|SHAP(z_i)|}{\|SHAP(z)\|_1}
    \label{eq:shapscore}
\end{align}
}
\noindent where $SHAP(z_i)$ is the SHAP score of the embedding feature $z_i$ and $SHAP(z)$ is a vector of the SHAP scores for all the embedding features in $z$.
In practice, we can compute SHAP scores for the embedding features by using a pretrained model that is trained on a publicly available dataset to avoid any extra privacy risks. Figure \ref{fig:noisescale - Hai} shows the distribution of sensitivity and importance scores across embedding features. 
% Some features are more sensitive and important than others, i.e., located within the top-right corner. Some features are either ``sensitive but not important'' (i.e., located within the top-left corner) or ``important but not sensitive'' (i.e., located within the bottom-right corner).

To capture the correlation between sensitivity and importance scores, we define a unifying score $\theta_i$ by a linear combination of $\alpha_i$ and $\beta_i$, as follows:
% such that it can be used to develop heterogeneous  randomization probabilities for embedding features, as follows:
{
\small
\begin{equation}
    \theta_i = \gamma\alpha_i + (1-\gamma)\big[\beta_{min} + (\beta_{max} - \beta_i)   
    \big] 
    \label{eq:overallscore} 
\end{equation}
}
where $\gamma \in [0, 1]$ is a weighted parameter to balance between the sensitivity score $\beta_i$ and the importance score $\alpha_i$, $\beta_{min} = \min_{j \in [d]}\beta_j$, $\beta_{max} = \max_{j \in [d]}\beta_j$, 
% the term $\beta_{min} + (\beta_{max} - \beta_i) \in [\beta_{min}, \beta_{max}]$. The inversion of $\beta_i$ ensures that the higher value of $\beta_i$ will have a smaller value in the inversion one.
The idea of Eq. \ref{eq:overallscore} is to separate between two set of features: the features have small values of $\alpha_i$ and high values of $\beta_i$ (top-left corner of Figure \ref{fig:noisescale - Hai}); and the features have high values of $\alpha_i$ and small values of $\beta_i$ (bottom-right corner in Figure \ref{fig:noisescale - Hai})
% more important (larger values of $\alpha_i$) and less sensitive (smaller value of $\beta_i$)  features (located within the bottom-right corner in Figure \ref{fig:noisescale - Hai}). 
% less important (smaller values of $\alpha_i$) and high sensitive (higher values of $\beta_i$) features (located within the top-left corner in Figure \ref{fig:noisescale - Hai}); and (2) more important (larger values of $\alpha_i$) and less sensitive (smaller value of $\beta_i$)  features (located within the bottom-right corner in Figure \ref{fig:noisescale - Hai}). 
For the other features, since they are both more (less) important and more (less) sensitive (located within the top-right and bottom-left corner in Figure \ref{fig:noisescale - Hai}), Eq. \ref{eq:overallscore} will smoothly combine the important and sensitive scores as a trade-off between importance and sensitiveness through the hyper-parameter $\gamma$ 

\paragraph{Randomizing Process} Given the unifying score $\theta_i$, we randomize the feature $i$ such that more important and less sensitive features (higher values of $\theta_i$) will have higher probabilities to stay the same, and vice versa.
That enables us to achieve better data utility. To achieve our goal, we assign a heterogeneous privacy budget $\varepsilon_i = \varepsilon_f\theta_i$ to the feature $i$. Then, we randomize each feature such that we provide $\varepsilon_i$-LDP for feature $i$ while addressing the privacy-utility trade-off. We tackle this by minimizing the difference between the original and randomized value of feature $i$ as follows. 

% The total privacy budget for the whole embedding feature vector ($\sum_{i=1}^d\varepsilon_i$) will be equal to $\varepsilon_f$ by following the composition theorem \cite{dwork14}: 
% $\sum_{i=1}^d\varepsilon_i = \varepsilon_f\sum_{i=1}^d\theta_i = \varepsilon_f$.

% This assignment will ensure that the features with higher values of $\theta_i$ will have higher values of $\varepsilon_i$, resulting in a higher chance that these features will be unchanged. Therefore, we can optimize the data utility since the more important and less sensitive features will have a higher chance to stay the same.

% \begin{figure}[t]
%     \centering\includegraphics[width=0.55\columnwidth]{images/rand_prob_new_eps_2.0_k_5.jpg} \vspace{-10pt}
%     \caption{Randomization probability distribution (Eq. \ref{eq:randprob}) of possible randomized values with $k=5$ bins and original value is $\frac{3}{5}$.} \vspace{-20pt}
%     \label{fig:randprob}
% \end{figure}

Without loss of generality, considering the value of each feature is in the domain $[0,1]$. To optimize the data utility, we design a randomizing process such that the randomized values should fall in the original domain and the values nearer to the original value will have higher sampling probability. 
However, by considering the continuous domain, the sampling probability of each point in the domain is minuscule, therefore, we discretize the domain $[0,1]$ by $k$ bins, resulting in a discrete domain $\{\frac{1}{k},\dots,1\}$. This discretizing process will limit the outcomes, leading to a higher probability for each value. Then, we transform the value of each embedding feature from the $[0,1]$ domain to the discrete domain. Formally, the value $z_i$ of embedding feature $i$ will have the value $\frac{t}{k}$ if $\frac{t-1}{k} \le z_i \le \frac{t}{k}, t \in \{1, 2, \dots, k\}$. This can done as a data preprocessing step without extra privacy risks. 

We randomize the value of each feature by the following rule: Given the value of the embedding feature $i$ is $z_i = \frac{t}{k}$, we randomize embedding feature $i$ such that it will have a randomized value $\frac{u}{k}, \forall u \in \{1, \dots, k\}$, with the probability 
{
\small
\begin{equation}
    Pr\Big(\frac{u}{k}\Big|\frac{t}{k}\Big) = \frac{1}{C_t}\exp\Big(-\frac{|u - t|}{k\sigma_i}\Big)
    \label{eq:randprob}
\end{equation}
}

\noindent where $\sigma_i$ is the noise scale of feature $i$ and $C_t$ is the normalization parameter and quantified by:
{
\small
\begin{equation}
    C_t = \sum_{u = 1}^k\exp\Big(-\frac{|u - t|}{k\sigma_i}\Big)
\end{equation}
}

\noindent Eq. \ref{eq:randprob} ensures that the values of $\frac{u}{k}$ that are closer to the original value $\frac{t}{k}$ will have a higher sampling probability, resulting in a closer distance of the original and randomized value of feature $i$ which optimizes the data utility.
% In Figure \ref{fig:randprob}, we illustrate the randomization probability distribution of all possible randomized values with $k=5$ and original value $\frac{3}{5}$ and this distribution is similar and general for all original values. 
To satisfy the assigned privacy budget $\varepsilon_i$ for feature $i$, we introduce the noise scale parameter $\sigma_i$, bounded to satisfy the guarantee of $\varepsilon_i$-LDP in section \ref{sec:privguar}.

Finally, we randomize each feature by applying \textsc{FeatureRR} on every features independently. The randomized features are concatenated together to create a randomized feature vector $\Tilde{z}$, which is used in the training process. The pseudo codes of \textsc{FeatureRR} is presented in Algorithm \ref{alg:featlvel} (Appendix \ref{appdix:algs} in our supplemental documents).

\subsection{Edge-aware Randomized Response (\textsc{EdgeRR})}

We present our \textsc{EdgeRR} mechanism in this section. First, we introduce \textsc{pHRG} model, an alternative to the HRG model. Second, we introduce our MCMC sampling process of \textsc{pHRG}, which leverages the public edges to optimize the utility of the \textsc{pHRG} while providing privacy protection for private edges

\paragraph{\textsc{pHRG} model} 

\begin{figure}[t]
% % \vskip 0.2in
    \centering     %%% not \center
    \subfigure[Graph $G(\mathcal{V}, \mathcal{E}_{pub} \cup \mathcal{E}_{pri})$]{\label{fig:a}\includegraphics[width=0.4\columnwidth]{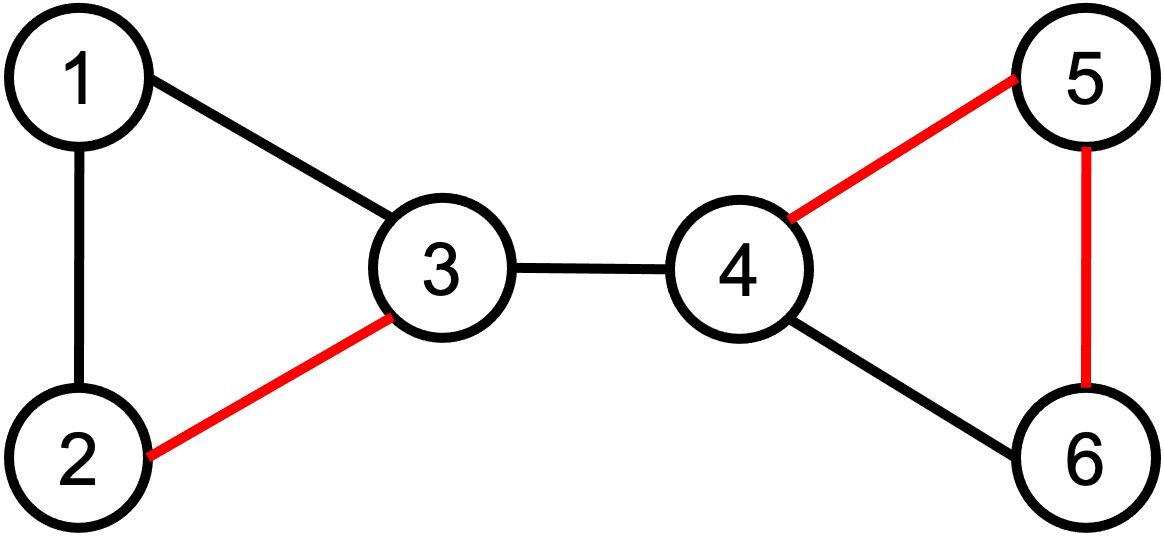}}
    \subfigure[HRG model ($\mathcal{D}$,$\{p_r, \Bar{p}_r\}$)]{\label{fig:b}\includegraphics[width=0.4\columnwidth]{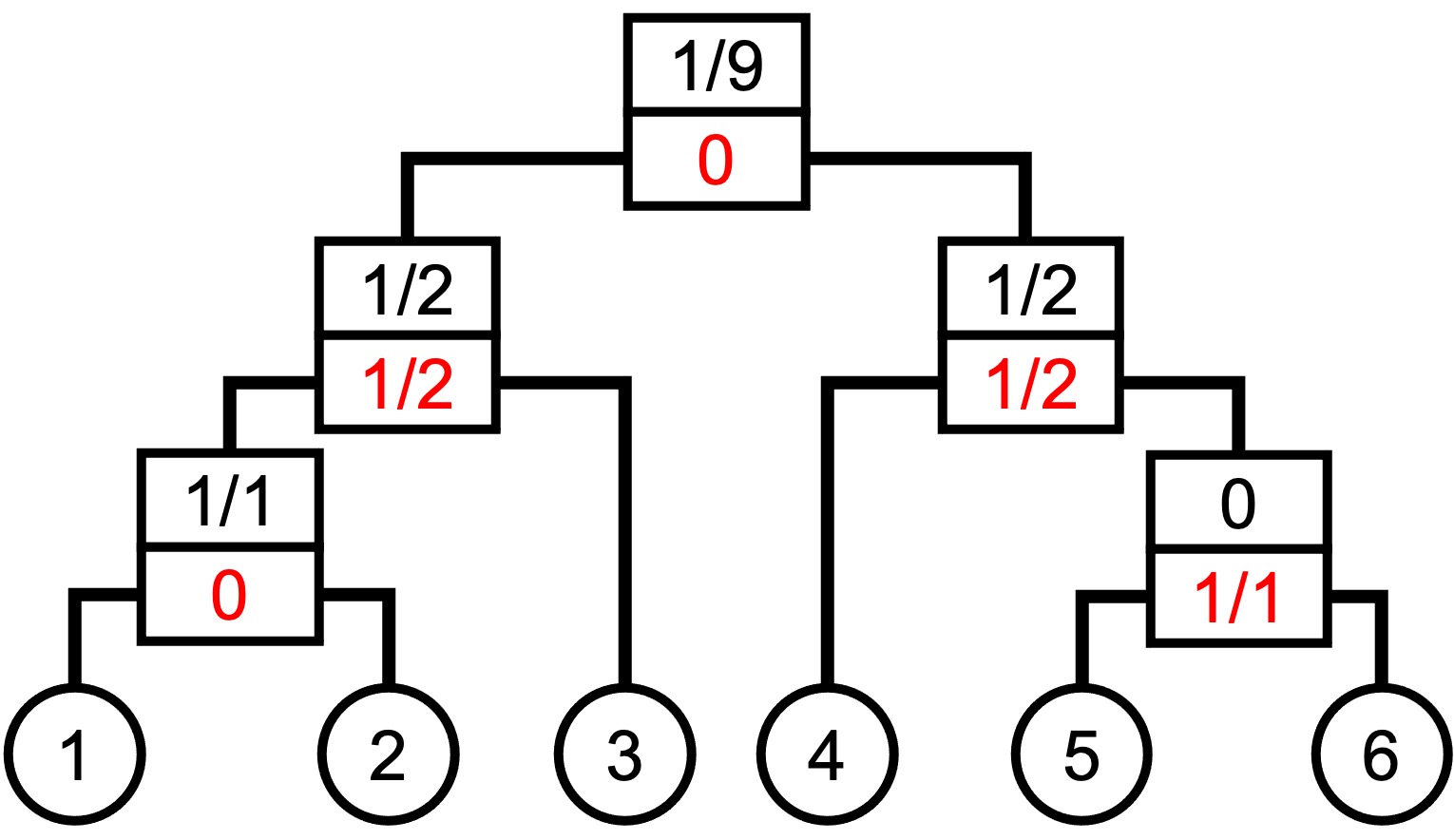}} \vspace{-5pt}
    \caption{An instance of HRG model. In (a), the red and black edges are the private and public edges, respectively. In (b), the black and red values are the values of $p_r$ and $\Bar{p}_r$ respectively.} 
    \vspace{-15pt}
    \label{fig:mhrg}
\end{figure}

\begin{figure}[t]
% % \vskip 0.2in
    \centering     %%% not \center
    \subfigure[$\mathcal{D}(r)$]{\label{fig:a}\includegraphics[width=0.2\columnwidth]{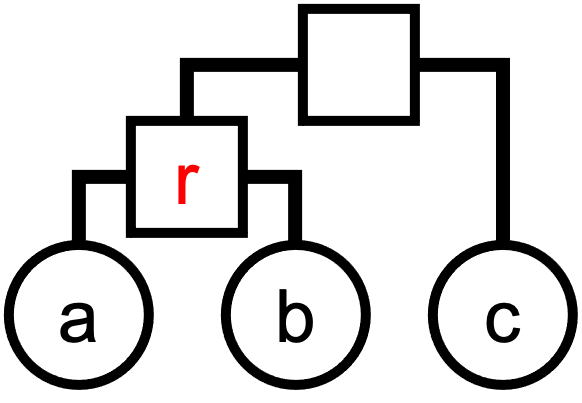}}
    \subfigure[Two possible versions of $\mathcal{D}'(r)$]{\label{fig:b}\includegraphics[width=0.45\columnwidth]{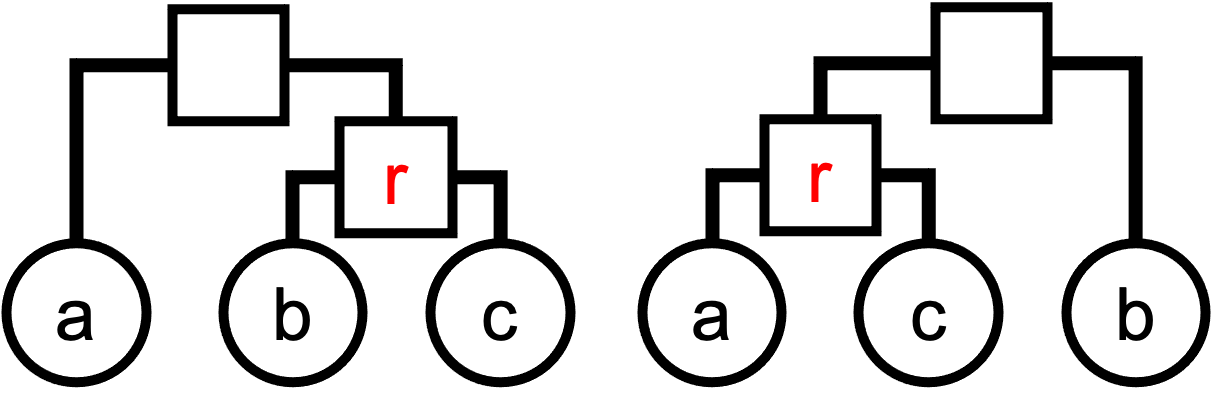}} \vspace{-5pt}
    \caption{Possible structures of subtree at an internal node $r$.}
    \label{fig:possiblesubtree}
    \vspace{-22pt}
    % \label{fig:hrg}
\end{figure}

Let us define a \textsc{pHRG} model (Figure \ref{fig:mhrg}) by a tuple $(\mathcal{D}, \{p_r, \Bar{p}_r\})$ where $\mathcal{D}$ is a dendrogram with the number of leaf nodes $n = |\mathcal{V}|$, the set of probabilities $\{p_r, \Bar{p}_r\}$ is associated with each internal node $r$ in $\mathcal{D}$. In this work, $p_r$ ($\Bar{p}_r$) is the probability of having a public (private) edge  between the leaf nodes from the left of $r$ and all the leaf nodes from the right of $r$. We define the likelihood of a dendrogram $\mathcal{D}$ as a product of the likelihood of public edges and the likelihood of private edges as follows:
{
\small
\begin{multline}
    \mathrm{L}(\mathcal{D}, \{p_r, \Bar{p}_r\}) = \mathrm{L}_{pub}(\mathcal{D}, \{p_r\})\mathrm{L}_{pri}(\mathcal{D}, \{\Bar{p}_r\}) \\
    = \prod_{r \in \mathcal{D}}p_r^{e_r}(1 - p_r)^{L_rR_r - e_r} 
    \times \Bar{p}_r^{\Bar{e}_r}(1 - \Bar{p}_r)^{\Bar{L}_r\Bar{R}_r - \Bar{e}_r} \label{eq:objfunc}
\end{multline}
}
\noindent where $e_r$ and $\Bar{e}_r$ are the public and private edges between the leaf nodes from the left of $r$ and leaf nodes from the right of $r$; $L_r$ and $R_r$ are the numbers of leaf nodes from the left and the right of $r$ respectively, such that each node has \textit{at least one public edge}; and $\bar{L}_r$ and $\bar{R}_r$ are the numbers of nodes from the left the right child of $r$ respectively, such that each node has \textit{at least one private edge}. In addition, by maximum likelihood, $p_r = \frac{e_r}{L_rR_r}$ and $\Bar{p}_r = \frac{\Bar{e}_r}{\Bar{L}_r\Bar{R}_r}$.

To find the most suitable \textsc{pHRG} model representing the graph $G$, we need to find the optimal dendrogram $\mathcal{D}^*$ that maximize the log-likelihood of Eq. \ref{eq:objfunc}, as follows: 
{
\small
\begin{align}
    \mathcal{D}^* & = \arg\max_{\mathcal{D}} \big(\mathbb{L}_{pub}(\mathcal{D}) + \mathbb{L}_{pri}(\mathcal{D})\big) \nonumber \\ 
    & = \arg\max_{\mathcal{D}} \big[-\sum_{r}N_r\chi(p_r) -\sum_{r}\Bar{N}_r\chi(\Bar{p}_r)\big]
\end{align}
}
\noindent where $\mathbb{L}_{pub}(\mathcal{D}) = \sum_{r}N_r\chi(p_r)$, $\mathbb{L}_{pri}(\mathcal{D}) = \sum_{r}\Bar{N}_r\chi(\Bar{p}_r)$, $\{r\}$ is the set of all internal nodes in $\mathcal{D}$, $N_r = L_rR_r$, $\Bar{N}_r = \Bar{L}_r\Bar{R}_r$, and $\chi(\tau) = \tau\log\tau + (1-\tau)\log(1-\tau)$.

% , and
% \begin{equation}
%     \mathbb{L} = \log\mathrm{L} = \mathbb{L}_{pub} + \mathbb{L}_{pri} = -\sum_{\{r\}}N_r\chi(p_r) -\sum_{\{r\}}\Bar{N}_r\chi(\Bar{p}_r)
%     \label{eq:logobjfunc}
% \end{equation}

\paragraph{Edge-level DP MCMC Sampling} It is expensive to find $\mathcal{D}^*$ by generating all $(2n-3)!!$  possible dendrograms. To address this problem, we propose a max-max MCMC process to approximate $\mathcal{D}^*$ while preserving the edge-level DP of $\mathcal{E}_{pri}$. Starting from a random dendrogram $\mathcal{D}_0$, each MCMC sampling step consist of two processes: (1) optimizing the dendrogram using the subgraph $G_{pub}(\mathcal{V}_{pub}, \mathcal{E}_{pub})$ and (2) optimizing the dendrogram using the subgraph $G_{pri}(\mathcal{V}_{pri}, \mathcal{E}_{pri})$ where $\mathcal{V}_{pub}$ ($\mathcal{V}_{pri}$) are the set of nodes in graph $G$ that has at least one public (private) edge. 
% Formally, \textsc{EdgeRR} optimizes the following surrogate objective function:
% In the first step, we optimize the maximization problem $\arg\max_{\mathcal{D}} \mathbb{L}_{pri}$ using only private edges $\mathcal{E}_{pri}$, their associated nodes $\mathcal{V}_{pri}$, and probabilities $\{\bar{p}_r\}$ , i.e., forming a subgraph of $G$, denoted $G_{pri}(\mathcal{V}_{pri}, \mathcal{E}_{pri})$. 
%This step can be conducted in an edge-level DP preserving MCMC process to protect $\mathcal{E}_{pri}$. However, the DP preserving noise can affect utility of the dendrogram, especially under a tight privacy budget. 
% In the second step, we use the resulting dendrogram $\mathcal{D}$ from the first step to optimize the maximization $\arg\max_{\mathcal{D}} \mathbb{L}_{pub}(\mathcal{D})$ in a MCMC process by using only public edges $\mathcal{E}_{pub}$, their associated nodes $\mathcal{V}_{pub}$, and probabilities $\{p_r\}$, forming a subgraph of $G$, denoted $G_{pub}(\mathcal{V}_{pub}, \mathcal{E}_{pub})$.
% That helps us to mitigate the data utility degradation caused by the first step. This optimization does not incur any privacy risk. Formally, \textsc{EdgeRR} optimizes the following surrogate objective function:
% {\small
% \begin{align}
%     \mathcal{D}^* = \arg\max_\mathcal{D}\Big[\mathbb{L}_{pub}\Big(\mathcal{D} \sim \max_{\mathcal{D}}\mathbb{L}_{pri}(\mathcal{D})\Big) + \\ \mathbb{L}_{pri}\Big(\mathcal{D} \sim \max_{\mathcal{D}}\mathbb{L}_{pub}(\mathcal{D})\Big)\Big] 
%     \label{minimax}
% \end{align}}

At a MCMC sampling step $t$, process (1) randomly samples a dendrogram $\mathcal{D}'$, and updates the current dendrogram $\mathcal{D}_t$ as:

{\small
\begin{align}
    \label{eq:dendrogramupdate}
    \mathcal{D}^{(1)}_t &= \left \{
    \begin{aligned}
        &\mathcal{D}', &&\text{with probability } \eta  \\
        &\mathcal{D}_{t-1}, &&\text{with probability } 1 - \eta
    \end{aligned} \right.
\end{align}}
\noindent where the acceptance probability $\eta = \min\Big(1, \frac{\exp{\mathbb{L}_{pub}(\mathcal{D}')}}{\exp{\mathbb{L}_{pub}(\mathcal{D}_{t-1})}}\Big)$. To sample $\mathcal{D}'$, we randomly choose an internal node $r$ (not the root) in $\mathcal{D}_{t-1}$ and randomly choose one of the two alternative possible structures of $r$ as $\mathcal{D}'$ (Figure \ref{fig:possiblesubtree}).

Similarly, at the (2) process \textsc{EdgeRR} randomly samples $\mathcal{D}'$ and updates the current dendrogram $\mathcal{D}_{t}$ as follows:
{
\small
\begin{align}
    \label{eq:dendrogramupdate}
    \mathcal{D}_{t} &= \left \{
    \begin{aligned}
        &\mathcal{D}', &&\text{with probability } \bar{\eta}  \\
        &\mathcal{D}^{(1)}_{t}, &&\text{with probability } 1 - \bar{\eta}
    \end{aligned} \right.
\end{align}
}
\noindent where the acceptance probability $\Bar{\eta}$ is computed as 
{
\small
\begin{equation}
\Bar{\eta} = \min\Big(1, \frac{\exp{(\frac{\varepsilon_{e1}}{\Delta_e}\mathbb{L}_{pri}\big(\mathcal{D}')\big)}}{\exp\big(\frac{\varepsilon_{e1}}{\Delta_e}\mathbb{L}_{pri}(\mathcal{D}_{t'-1})\big)}\Big)
\label{Laplace Dendrogram}
\end{equation}
}
\noindent with $\Delta_e$ is the global sensitivity of $\mathbb{L}_{pri}(\cdot)$ bounded in Lemma \ref{lemma:deltaE} and a privacy budget $\varepsilon_{e1}$. Since, the MCMC sampling process is reversible and ergodic \cite{kddbaseline}, there exists only one equilibrium state, which assures the convergence condition.

After the sampling process, given $(\mathcal{D}^*, \{p_r,\Bar{p}_r\})$, we employ  \verb|CalculateNoisyProb| algorithm (Algorithm \ref{alg:CalculateNoisyProb} in Appendix \ref{appdix:algs}) \cite{kddbaseline} to add Laplacian noise to $\{\Bar{p}_r\}$ with a privacy budget $\varepsilon_{e2}$.
We use the perturbed dendrogram to generate the edge-level DP-preserving subgraph $\Tilde{G}_{pri}$, then, we merge $\Tilde{G}_{pri}$ with the public graph $G_{pub}$ to construct the (complete) edge-level DP-preserving graph $\Tilde{G} = G_{pub} \cup \Tilde{G}_{priv}$. The graph $\Tilde{G}$ will be used to train the GNNs as the replacement for $G$ to preserve the privacy of private edges $\mathcal{E}_{pri}$. The pseudo codes of \textsc{EdgeRR} is presented in Algorithm \ref{alg:edgelevel} (Appendix \ref{appdix:algs}).
\section{Privacy Guarantees}
\label{sec:privguar}

This section analyzes the privacy guarantee of \textsc{HeteroRR} at the embedding feature-level LDP and the edge-level DP. 

% In fact, we need to find 1) a closed-form bound of $\mu$ in Eq. \ref{eq:bitawareprob} to achieve $\varepsilon_f$-LDP-preserving $\tilde{B}$ given the bit-string $B$ and 2) a closed-form bound of $\Delta_e$ in Eq. \ref{Laplace Dendrogram} to preserve $\varepsilon_{e1}$ edge-level DP.

\textit{Bounding $\sigma_i$ in \textsc{FeatureRR}:} To satisfy $\varepsilon_f$-LDP for the embedding feature $z$, we need to bound the noise scale $\sigma_i$ to assure $\varepsilon$-LDP for the embedding feature $z$.
\begin{theorem}
    \label{theo:featureprivacy}
    For each feature $i \in [d]$, if $\sigma_i \ge \frac{(k - 1)}{k\epsilon\theta_i}$, then our mechanism preserves $\epsilon$-LDP for the whole feature vector $x$.
\end{theorem}
\noindent The proof of Theorem \ref{theo:featureprivacy} is in the Appendix \ref{proof:featureprivacy} of our supplemental documents.

\textit{Edge-level DP:}
First, we bound the global sensitivity $\Delta_e$. 
\begin{lemma}
    $\Delta_e$ monotonically increases as $n \rightarrow +\infty$, and
    {
    \small
    \begin{equation}
        \Delta_e = \log N_{max} + \log\Big(1+\frac{1}{N_{max}-1}\Big)
    \end{equation}
    }
    where $N_{max} = \frac{|\Bar{\mathcal{V}}|^2}{4}$ when $|\Bar{\mathcal{V}}|$ is even and $N_{max} = \frac{|\Bar{\mathcal{V}}|^2-1}{4}$ when $|\Bar{\mathcal{V}}|$ is odd with $\Bar{\mathcal{V}} \subset \mathcal{V}$ is the set of private nodes. 
    \label{lemma:deltaE}
\end{lemma}
\noindent The proof of Lemma \ref{lemma:deltaE} is in the Appendix \ref{proof:deltaE} of our supplemental documents. Given the bounded sensitivity $\Delta_e$ in Lemma \ref{lemma:deltaE}, our HRG sampling process satisfies edge-level $\varepsilon_{e1}$-DP, since the exponential mechanism has been proved satisfying the DP constraint with a desired privacy budget $\varepsilon_{e1}$ \cite{dwork14}.

Regarding the \verb|CalculateNoisyProb| algorithm, for two neighboring graphs $G$ and $G'$ that are different at one private edge, the global sensitivity is $1$. Thus, perturbing the sampled dendrogram $\mathcal{D}^*$ satisfies $\varepsilon_{e2}$-DP to protect private edges. Following the sequential composition theorem \cite{dwork14}, \textsc{EdgeRR} satisfies $\varepsilon_e$-DP with $\varepsilon_e = \varepsilon_{e1} +\varepsilon_{e2}$ to protect private edges. 

% \begin{figure}
%     \centering
%     \subfigure[Randomization probability $q_i$ with respect to the overall score $\theta_i$]{\label{fig:randprobthet}\includegraphics[scale=0.22]{images/randomize_probability_with_theta.jpg}}
%     \hfill
%     \subfigure[Expected error bound with respect to the overall score $\theta_i$]{\label{fig:EERthet}\includegraphics[scale=0.22]{images/error_with_theta.jpg}} \vspace{-10pt}
%     \caption{The effect of the overall score toward the randomization probability and the expected error bound with different $d$ and $\varepsilon_f = 0.01$}
% \end{figure}

\section{Experimental Results}

\begin{figure}[t]
% % \vskip 0.2in
    \centering     %%% not \center
    \subfigure[FLICKR-MIR]{\label{fig:mir_feat}\includegraphics[width=0.49\columnwidth]{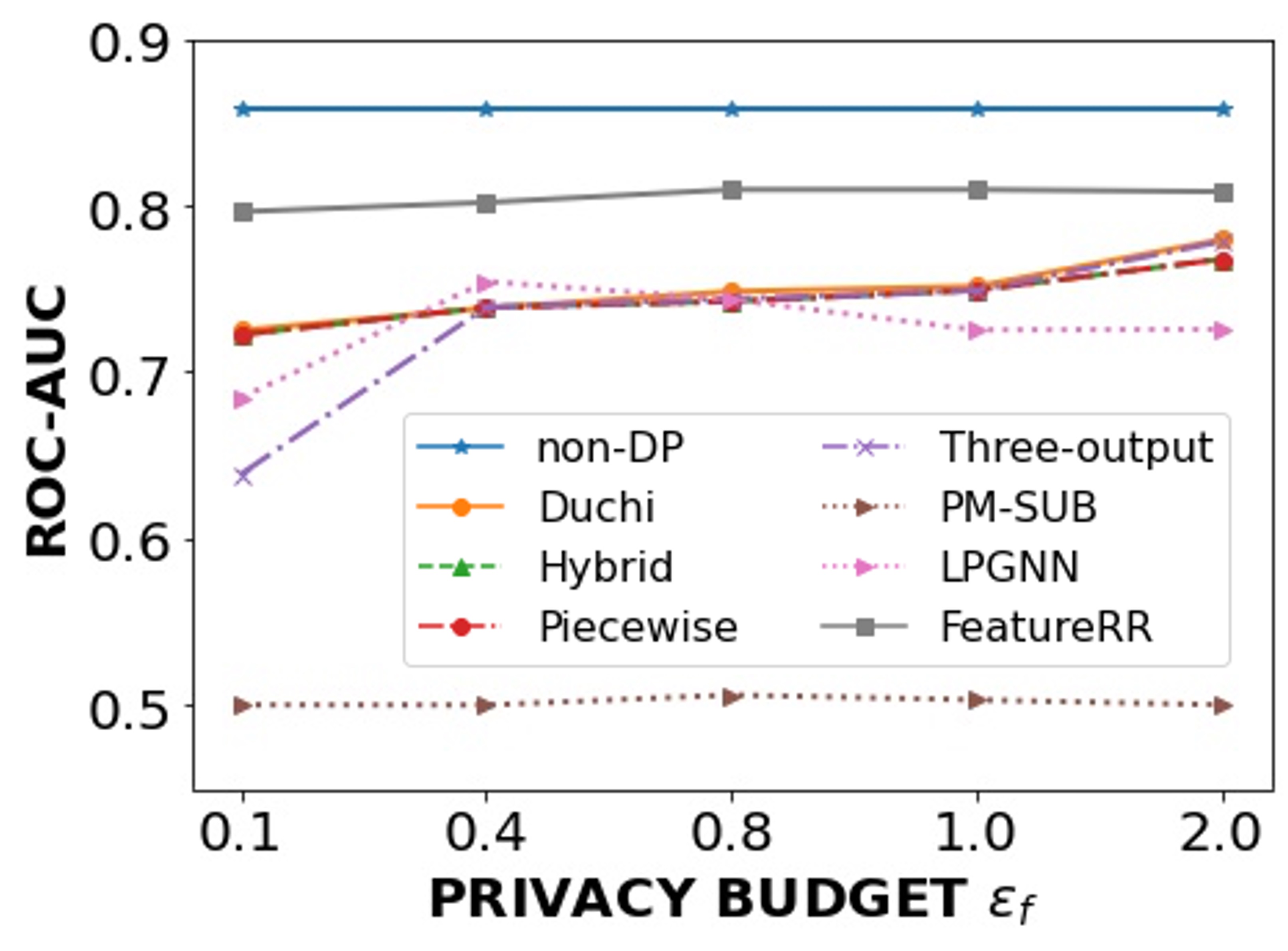}}
    \hfill
    \subfigure[PPI]{\label{fig:ppi_feat}\includegraphics[width=0.49\columnwidth]{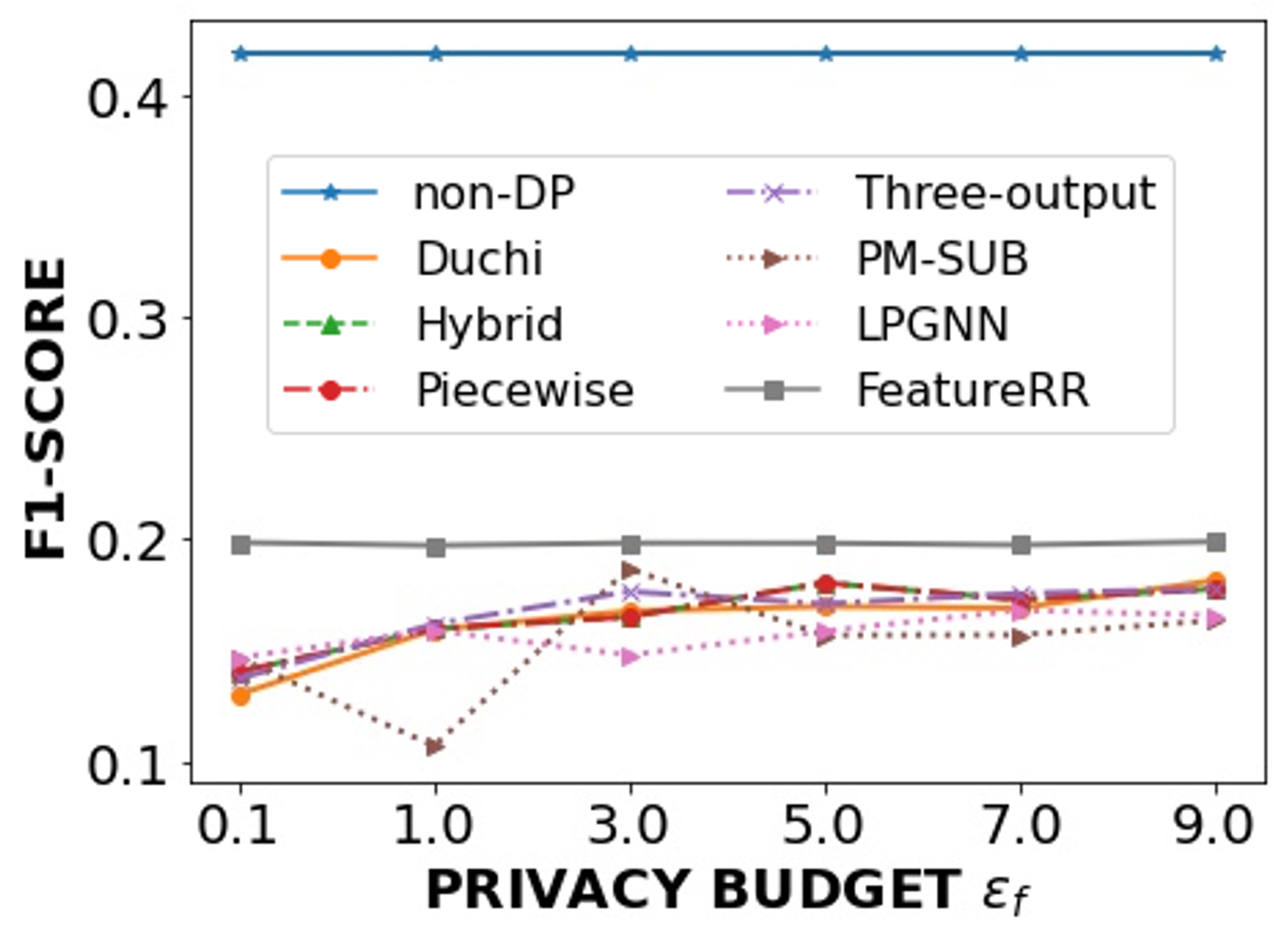}}   \vspace{-10pt}
    \caption{Model performance of the feature-level protection.} 
    \vspace{-20pt}
    \label{fig:resBRand}
\end{figure}

We conduct extensive experiments on benchmark datasets to illustrate interplay between privacy budget and model utility in \textsc{HeteroRR} at the feature level, the edge level, and how well it can defend against the PIAs at both levels.

% \subsection{Datasets, Models and Evaluation metrics}

\paragraph{Datasets, Model and Metrics} 

We consider two datasets including FLICKR-MIR and PPI \cite{graphsage}. For the PPI dataset, we select the proportion of private edges $\rho \in \{0.05, 0.1, 0.2\}$ to construct the set of private edges in each graph. For the FLICKR-MIR dataset, we consider the edges constructed by ``taken by friends'' private ($\rho \approx 0.305$) and others are public edges. We use ResNet-50 \cite{Resnet} with ImageNet weights and RetinaFace \cite{retinaface} to extract embedding features and faces from the images. 

We employ the Graph Convolutional Network (GCN) \cite{GCN} in every experiment. We use an average aggregation function for each layer in the GCN models as in \cite{GCN}. We use F1-score to evaluate the model performance on the PPI dataset, and we use \textit{ROC-AUC} metric to evaluate the model performance on the FLICKR-MIR dataset. All statistical tests are 2-tail t-tests.

\paragraph{Baselines}
We consider well-known state-of-the-art LDP mechanism baselines for the feature-level privacy protection, including Duchi mechanism \cite{duchimechanism}, Piecewise mechanism \cite{piecewismechanism}, Hybrid mechanism \cite{piecewismechanism}, Three-output mechanism \cite{threeoutputmechanism}, Sub-optimal mechanism \cite{threeoutputmechanism}, and (LPGNN) \cite{locallyprivateGNN}. Regarding the edge-level, we consider three baselines: privHRG \cite{kddbaseline}, EdgeRand \cite{linkTeller}, and LapGraph \cite{linkTeller}. We include the clean model (non-DP) to show the upper bound of the model performance.

% To evaluate our \textsc{HeteroRR}, we consider the following three settings: \textbf{(1)} We randomize nodes' embedding features and train the GCN model with the non-DP setting for the graphs' structure (i.e., $\varepsilon_e = \infty$); \textbf{(2)} We randomize graphs' structure and train the GCN model with the non-DP setting for nodes' embedding features (i.e., $\varepsilon_f = \infty$); \textbf{(3)} We apply \textsc{FeatureRR} on each node's embedding features, \textsc{EdgeRR} on each graph, and combine the randomized nodes' features with the DP-preserving graph to train the GCN model; and \textbf{(4)} We test \textsc{HeteroRR} against PIAs.

\paragraph{Results on the Embedding Feature-Level} 

Figures \ref{fig:resBRand} illustrates the GCN performance associated with the change of $\varepsilon_f$. 
% We consider a wide range of rigorous privacy budget $\varepsilon_f \in [0.1, 2.0]$. 
\textsc{FeatureRR} achieves the best model performance compared to the baselines in all three datasets. For the FLICKR-MIR dataset, \textsc{FeatureRR} has $7.6\%$ improvements on average, respectively, i.e., $p\text{-value} = 3.0e-05$, compared with the best baseline (LPGNN). In the PPI dataset, 
% due to the significant drop of GCN performance in the extremely high-privacy domain, we use bigger privacy budgets $\varepsilon_f \in [0.1, 9.0]$. 
\textsc{FeatureRR} improves the model performance $2.6\%$ compared with the best baseline, i.e., the Three-output mechanism, with $p\text{-value} = 0.0031$. 
% The gap between \textsc{FeatureRR} and the baselines is wider with bigger $\varepsilon_f$.
% The results are concise since \textsc{FeatureRR} adjusts the randomization probability toward the importance of the features, which assures the models still have rich separable information of the nodes.

\paragraph{Results on the Edge-level}

\begin{figure}[t]
% \vspace{-10pt}
    \centering 
    \subfigure[FLICKR-MIR]{\label{fig:mir_edge}\includegraphics[width=0.49\columnwidth]{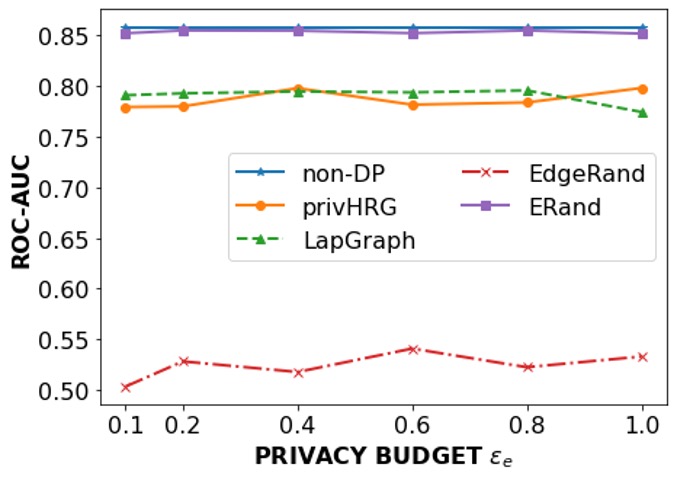}} 
    \subfigure[PPI]{\label{fig:ppi_edge}\includegraphics[width=0.49\columnwidth]{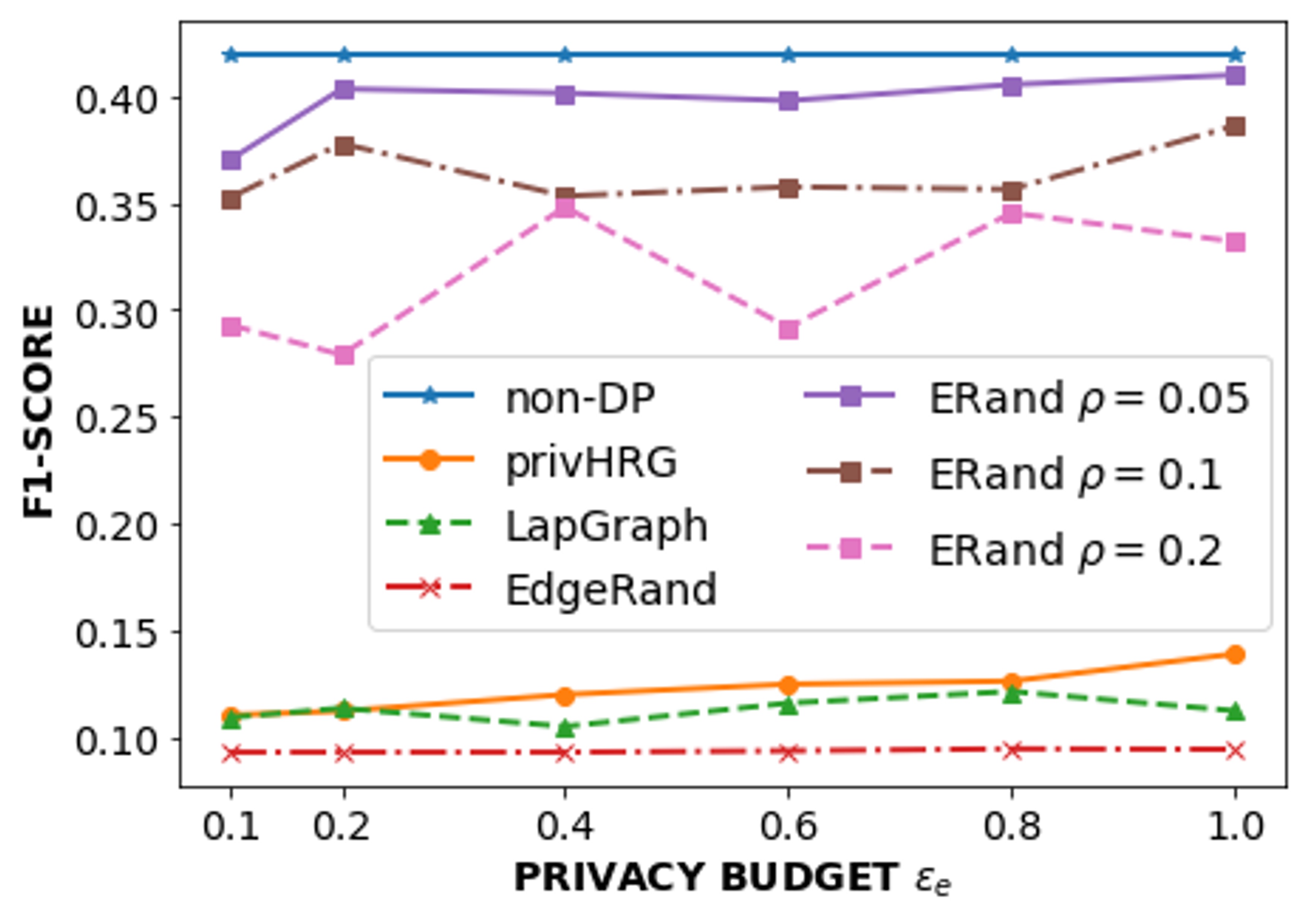}}
    \vspace{-5pt}
    {
    \footnotesize
    \caption{Model performance of the edge-level protection.} \label{fig:resERand}
    }
    \vspace{-10pt}
    
\end{figure}

Figures \ref{fig:resERand} illustrate the results on the FLICKR-MIR, and PII datasets. 
% We consider a wide range of $\varepsilon_e \in [0.1, 1.0]$. 
In the FLICKR-MIR, \textsc{EdgeRR} outperforms all the baselines where textsc{EdgeRR} improves $6.3\%$ over the best baseline (LapGraph) with $p\text{-value} = 3.6e-09$. 
% and has the model performance close to the clean model (non-DP) with a tiny drop in the ROC-AUC. \
% Since the \textsc{EdgeRR} leverages the heterogeneity in the graph structure, \textsc{EdgeRR} preserved high graph structural utility results in high models performance.
In the PPI dataset (Figure \ref{fig:ppi_edge}), we compare the model performance of each algorithm given different values of $\rho$. We observe that the F1-score of GCN trained is higher when $\rho$ is smaller since the lower $\rho$ reduces the number of private edges being randomized. 
% with smaller $\rho = 0.05$ is averagely $9.4\%$ higher than the GCN trained with $\rho = 0.1$, i.e., $p\text{-value} = 0.002$. Moreover, the F1-score of GCN trained with $\rho = 0.1$ is averagely $16.5\%$ higher than the GCN trained with $\rho = 0.2$, i.e., $p\text{-value} = 0.005$. This is straightforward since the lower $\rho$ reduces the number of private edges being randomized. 

% Comparing with the best baseline (privHRG), \textsc{EdgeRR} trained on the PPI dataset with $\rho = 0.2$ still significantly improve the model performance by $19.2\%$ with $p\text{-value} = 4.35e-08$. Since the \textsc{EdgeRR} only randomizes the set of private edges, and the randomization of \textsc{EdgeRR} is optimized by leveraging the public edges, the DP-preserving graph resulted by \textsc{EdgeRR} preserved high graph structural utility of the training graph results in high models performance.

\paragraph{Defending against PIAs}

We conduct the LinkTeller \cite{linkTeller} and Image Reconstruction \cite{cvpr2016} attack to the GNNs under the protection of \textsc{HeteroRR} to test its defensive power. 
% LinkTeller is the state-of-the-art attack to infer edges from GCNs and the image reconstruction attack \cite{cvpr2016} is highly efficient to reconstruct images from embedding features \cite{cvpr2016} which matches our threat model.

\textbf{Defending the LinkTeller \cite{linkTeller}.}
We evaluate \textsc{EdgeRR} against the LinkTeller attack to analyze its ability in protecting private edges.
For the non-DP model, the LinkTeller shows an efficient attack performance (ROC-AUC is 0.91) on the FLICKR-MIR dataset. When we apply \textsc{EdgeRR} to protect the private edges, the performance of the LinkTeller is significantly reduced to nearly random guess $[0.52, 0.54]$ given a wide range of the privacy budget $\varepsilon_e \in [0.1, 1.0]$. Therefore, \textsc{EdgeRR} is effective in defending the LinkTeller attack.

\begin{figure}[t]
    \centering
    \includegraphics[width=0.8\columnwidth]{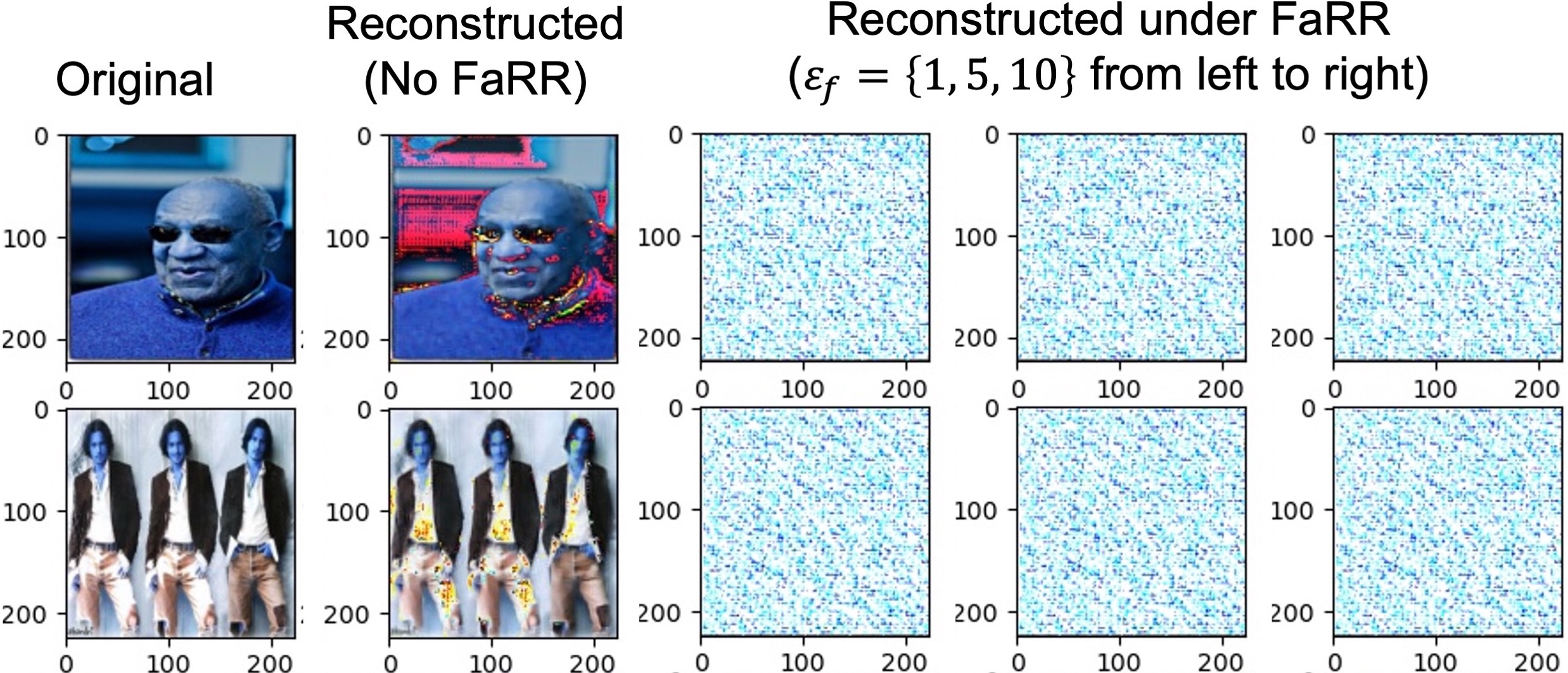} \vspace{-10pt}
    \caption{Results of \cite{cvpr2016} attack reconstructing the raw images from embedding features extracted from the layer $7^{\text{th}}$ over 176 layers of ResNet-50} \vspace{-20pt}
    \label{fig:ReconstructAttack}
\end{figure}

\textbf{Defending image reconstruction attack \cite{cvpr2016}.} 
% Since no complete work studies the feature inference attack of GNNs, we test our \textsc{FeatureRR} on the image reconstruction attack \cite{cvpr2016}. 
We train an attacker on the ImageNet dataset to reconstruct the images from embedding features extracted from the pre-trained ResNet-50 model; 
% \cite{Resnet}; 
then, we use the trained attacker to reconstruct the image from the embeddings 
% of the FLICKR-MIR datasets. The attacker is successful if it can infer faces from reconstructed images.  
The results of the reconstructed and original images are in Figure \ref{fig:ReconstructAttack}. We found that \textsc{FeatureRR} successfully prevent the attacks due to the power of LDP which is consistent with the previous studies.

\section{Conclusion}

In this paper, we present \textsc{HeteroRR}, a mechanism to simultaneously protect nodes' embedding features and private edges under LDP and DP protections in training GNNs. By balancing the sensitivity and importance of features and edges \textsc{HeteroRR} retains high data and model utility under the same privacy protection in training GNNs compared with existing baseline approaches. Also, our \textsc{HeteroRR} is resistant to PIAs, such as LinkTeller and image reconstruction attacks.

\section*{Acknowledgement}
This work is partially supported by grants NSF IIS-2041096, NSF CNS-1935928/1935923, NSF CNS-1850094, and unrestricted gifts from Adobe System Inc.

\bibliography{main}
\bibliographystyle{IEEEtran}

\appendices
\onecolumn

\section{Proof of Theorem
\ref{theo:featureprivacy}}
\label{proof:featureprivacy}
\begin{proof}
    Considering an arbitrary feature $i$ with noise scale $\sigma_i$ and score $\theta_i$. 
    To satisfy $\varepsilon_i$-LDP, for any pair of input value $t/k, t'/k \in \{\frac{1}{k}, \dots, 1\}$,  we bound the following proportion:
    
    \begin{align}
       & \frac{Pr(u/k|t/k)}{Pr(u/k|t'/k)} \le \argmax_{u,t,t'}\frac{Pr(u/k|t/k)}{Pr(u/k|t'/k)} = \frac{Pr(1/k|1/k)}{Pr(1/k|k/k)} \nonumber \\
       &= \exp[\frac{|1/k-1/k| - |k/k-1/k|}{\sigma_i}] \le \exp\Big(\frac{k-1}{k\sigma_i}\Big) \le e^{\varepsilon_i} \nonumber
    \end{align}
    
    Therefore, we have:
    \begin{equation}
        \sigma_i \ge \frac{(k - 1)}{k\varepsilon_i}
    \end{equation}
        
    By the strong composition theorem, the total privacy budget for the whole feature vector is calculated as follows:
        \begin{align}
            \sum_{i = 1}^d\varepsilon_i = \varepsilon_f\sum_{i = 1}^d\theta_i = \varepsilon_f
        \end{align}
    which completes our proof.
\end{proof}

\section{Proof of Lemma \ref{lemma:deltaE}}
\label{proof:deltaE}
\begin{proof}
Considering two neighbor graphs $G(\mathcal{V}, \mathcal{E}_{pub}\cup\mathcal{E}_{pri})$, $G(\mathcal{V}, \mathcal{E}_{pub}\cup\mathcal{E}'_{pri})$ which are different at one private edge. Without loss of generality, we consider $|\mathcal{E}_{pri}| = |\mathcal{E}'_{pri}|-1$. First of all, we fix $\Bar{N}_r$ and let $f(\Bar{e}) = \chi(\Bar{p}_r) - \chi(\Bar{p}'_r)$ and $\Delta_e = \text{max}|f(\Bar{e})|$. The second order derivative of $h(\Bar{p}_r)$ is given by: $\chi ''(\Bar{p}_r) = -\frac{1}{1 - \Bar{p}} - \frac{1}{\Bar{p}} < 0, \forall \Bar{p}$. Therefore, $\chi '(\Bar{p}_r)$ is a monotonically decreasing function which implies that $f(\Bar{e})$ is also a monotonically decreasing function ($\Bar{p}_r > \Bar{p}_r'$). With the monotonic property of $f(\Bar{e})$ we can derive value of $\Delta_e$ when $\Bar{e} = 1$ or $\Bar{e} = \Bar{N}_{max}$. Fixing $\Bar{e}_r = 1$ and vary $\Bar{N}_r$, therefore, $\Delta_e = \max_{\Bar{N}_r}|f(\Bar{N}_r)|$, where
\begin{equation}
 f(\Bar{N}_r) = 1\log\frac{1}{\Bar{N}_r} + (N-1)\log\Big(1 - \frac{1}{\Bar{N}_r}\Big)
\end{equation}

The first order derivative of $f(\Bar{N}_r) = \log\Big(1 - \frac{1}{\Bar{N}_r}\Big) < 0, \forall \Bar{N}_r$, so that $f(\Bar{N}_r)$ is monotonically decreasing by $\Bar{N}_r$. Since $f(\Bar{N}_r) < 0, \forall \Bar{N}_r \in [1, +\infty]$,  we conclude that $\Delta_e = -\min(f(\Bar{N}_r)) = -f(\Bar{N}_{max})$. Therefore,
\begin{equation}
\begin{aligned}
    \Delta_e &= \log(\Bar{N}_{max}) - (\Bar{N}_{max}-1)\log\Big(\frac{\Bar{N}_{max} - 1}{\Bar{N}_{max}}\Big) \\
            &= \log(\Bar{N}_{max}) - \log\Big(\frac{\Bar{N}_{max} - 1}{\Bar{N}_{max}}\Big)^{(\Bar{N}_{max}-1)} \\
            &= \log(\Bar{N}_{max}) - \log\Big(1 - \frac{1}{\Bar{N}_{max}}\Big)^{(\Bar{N}_{max}-1)}
\end{aligned}
\end{equation}

According to Cauchy's inequality, $N_{max} \le \frac{(\Bar{L}_r + \Bar{R}_r)^2}{4}$ and the equality happens when $\Bar{L}_r = \Bar{R}_r = \frac{|\Bar{V}|}{2} \rightarrow N_{max} = \frac{|\Bar{V}|^2}{4}$ when $|\Bar{V}|$ is even and $N_{max} = \frac{|\Bar{V}|^2-1}{4}$ when $|\Bar{V}|$ is odd.
\end{proof}

\newpage
\section{Algorithms}
\label{appdix:algs}
\begin{algorithm}[h]
\label{alg:featurelevel}
\footnotesize
\caption{Feature level protection (\textsc{FeatureRR})}\label{alg:featlvel}
\begin{algorithmic}[1]
\STATE \textbf{Input}: Privacy budget $\varepsilon_f$, number of bins $k$, number of features $d$, embedding features $z$, importance score vector $\alpha$, sensitivity score vector $\beta$, hyper-parameter $\gamma$.
\STATE \textbf{Output}: Perturbed embedding features $\Tilde{z}$
\STATE Normalize the values of $z$ to $[0,1]$
\STATE $\theta_i \leftarrow \gamma\alpha_i + (1-\gamma)\big[\beta_{min} + (\beta_{max} - \beta_i)\big], \forall i \in [d]$
\STATE $\theta_i \leftarrow  \frac{\theta_i}{\sum_{j = 1}^d\theta_j}, \forall i \in [d]$
\STATE $\varepsilon_i \leftarrow \theta_i\varepsilon_f \mbox{ and } \sigma_i = \frac{k-1}{k\varepsilon_i}, \forall i \in [d]$
\FOR {$i \in \{1, \dots, d\}$}
    \FOR {$t \in \{1, \dots, k\}$}
        \STATE if $\frac{t-1}{k} \le z_i \le \frac{t}{k}$: $z_i \leftarrow \frac{t}{k}$
    \ENDFOR 
\ENDFOR 
\STATE Initialize empty vector $\tilde{z}$ same shape with $z$.
\FOR {$i \in \{1, \dots, d\}$}
    \STATE $\Tilde{z}_i = \begin{cases} \frac{1}{k}, & \textit{with probability } Pr(\frac{1}{k} |z_i) \\ 
    $\dots$ \\
    \frac{k}{k}, & \textit{with probability } Pr(\frac{k}{k} |z_i) \end{cases}$
\ENDFOR
\STATE return $\Tilde{z}$
\end{algorithmic} 
\end{algorithm}

\begin{algorithm}[h] 
\label{alg:ERand}
\footnotesize
\caption{Edge level protection \textsc{EdgeRR}}\label{alg:edgelevel}
\begin{algorithmic}[1]
\STATE \textbf{Input}: Input graph ${G}(\mathcal{V}, \mathcal{E}_{pub} \cup \mathcal{E}_{pri})$, privacy budget $\varepsilon_{e1}$.
\STATE \textbf{Output}: randomized graph $\tilde{G}(\mathcal{V}, \mathcal{E}_{pub} \cup \tilde{\mathcal{E}}_{pri})$
\STATE Initialize the MCMC by a random dendrogram $D_o$;
\STATE $A \leftarrow$ the adjacency matrix of $G$
\FOR {each step $t$ of MCMC}
    \FOR{each public step}
        \STATE Randomly pick an internal node r in $D_{t-1}$
        \STATE Sample $\mathcal{D}'$ from a possible structure of substree at $r$
        \STATE Accept $\mathcal{D}_t$ = $D'$ with propability $\min\Big(1, \frac{\exp{\mathbb{L}_{pub}(\mathcal{D}')}}{\exp{\mathbb{L}_{pub}(\mathcal{D}_{t-1})}}\Big)$
    \ENDFOR 
    \FOR{each private step}
        \STATE Randomly pick an internal node $r$ in $D_{t-1}$
        \STATE Sample $\mathcal{D}'$ from a possible structure of substree at $r$
        \STATE Accept $\mathcal{D}_t$ = $D'$ with propability $\min\Big(1, \frac{\exp{(\frac{\varepsilon_{e1}}{\Delta_e}\mathbb{L}_{pri}\big(\mathcal{D}')\big)}}{\exp\big(\frac{\varepsilon_{e1}}{\Delta_e}\mathbb{L}_{pri}(\mathcal{D}_{t'-1})\big)}\Big)$
    \ENDFOR 
    \IF{converged}
        \STATE $\mathcal{D}* = \mathcal{D}_t$
    \ENDIF
\ENDFOR 
\STATE Applying \textbf{CalculateNoisyProb($G_{pri}(\mathcal{V}_{pri}, \mathcal{E}_{pri}),\mathcal{D}*,\varepsilon_{e2},r_{root}$)}

\STATE Initialize $\tilde{A}$ be zeros matrix
\FOR{each pair of $u,v \in \mathcal{V}$}
    \IF{$u \in \mathcal{V}_{pri}$ and $v \in \mathcal{V}_{pri}$}
        \STATE Find the lowest common ancestor $r$ of $u$ and $v$
        \STATE $\tilde{A}_{u,v} \sim Ber(1,\tilde{p}_r)$ (Bernoulli's distribution); $\tilde{A}_{v,u} =\tilde{A}_{u,v}$
    \ELSE 
        \STATE $\tilde{A}_{u,v} = A_{u,v}$; $\tilde{A}_{v,u} =\tilde{A}_{u,v}$
    \ENDIF
\ENDFOR
\STATE Construct $\tilde{G}$ from $\tilde{A}$ and return $\tilde{G}$
\end{algorithmic} 
\end{algorithm} 

\begin{algorithm}[h]
\footnotesize
\caption{\texttt{CalculateNoisyProb}($G_{pri}, \mathcal{D}*, \varepsilon_{e2}, r$) \cite{kddbaseline}}\label{alg:CalculateNoisyProb}
\begin{algorithmic}[1]
\STATE \textbf{Input}: Private graph $G_{pri}(\mathcal{V}_{pri}, \mathcal{E}_{pri})$, sampled dendrogram $\mathcal{D}*$, privacy budget $\varepsilon_{e2}$, internal node $r$
\STATE \textbf{Output}: Perturbed set $\{\Tilde{p}_r\}$

\STATE $\lambda_b \leftarrow \frac{1}{\varepsilon_{e2}\Bar{L}_r\Bar{R}_r}$; $\lambda_c \leftarrow \frac{1}{\varepsilon_{e2}(\Bar{L}_r+\Bar{R}_r)(\Bar{L}_r+\Bar{R}_r-1)}$
\IF {$\lambda_b \ge \tau_1$ and $\lambda_c \ge \tau_e$}
    \STATE $\Tilde{e}_r \leftarrow$ number of edges in the sub-graph induced from all private leave nodes of sub-tree rooted at $r$;
    \STATE $\Tilde{p}_r \leftarrow \min\Big(1, \frac{\Tilde{e}_r + Lap(\frac{1}{\varepsilon_{e2}})}{(\Bar{L}_r + \Bar{R}_r)(\Bar{L}_r + \Bar{R}_r -1)/2}\Big)$
    \FOR {each child internal node $r'$ of $r$}
        \STATE $\Tilde{p}_r' \leftarrow \min\Big(1, \frac{\Tilde{e}_r + Lap(\frac{1}{\varepsilon_{e2}})}{(\Bar{L}_r + \Bar{R}_r)(\Bar{L}_r + \Bar{R}_r -1)/2}\Big)$
    \ENDFOR 
\ELSE
     \STATE $\Tilde{p}_r \leftarrow \min\Big(1, \frac{\Bar{e}_r + Lap(\frac{1}{\varepsilon_{e2}})}{\Bar{L}_r\Bar{R}_r}\Big)$
     \STATE $r_L \leftarrow \text{left child of }r$; $r_R \leftarrow \text{right child of }r$;
     \STATE \texttt{CalculateNoisyProb}($G_{pri}, \mathcal{D}*, \varepsilon_{e2}, r_L$)
     \STATE \texttt{CalculateNoisyProb}($G_{pri}, \mathcal{D}*, \varepsilon_{e2}, r_R$)
\ENDIF
\STATE return $\{\Tilde{p}_r\}$
\end{algorithmic} 
\end{algorithm}

\newpage
\onecolumn
\section{Supplementary results}
\label{appdx:suppres}

\begin{figure}[h]
% % \vskip 0.2in
    \centering
    \includegraphics[width=0.9\columnwidth]{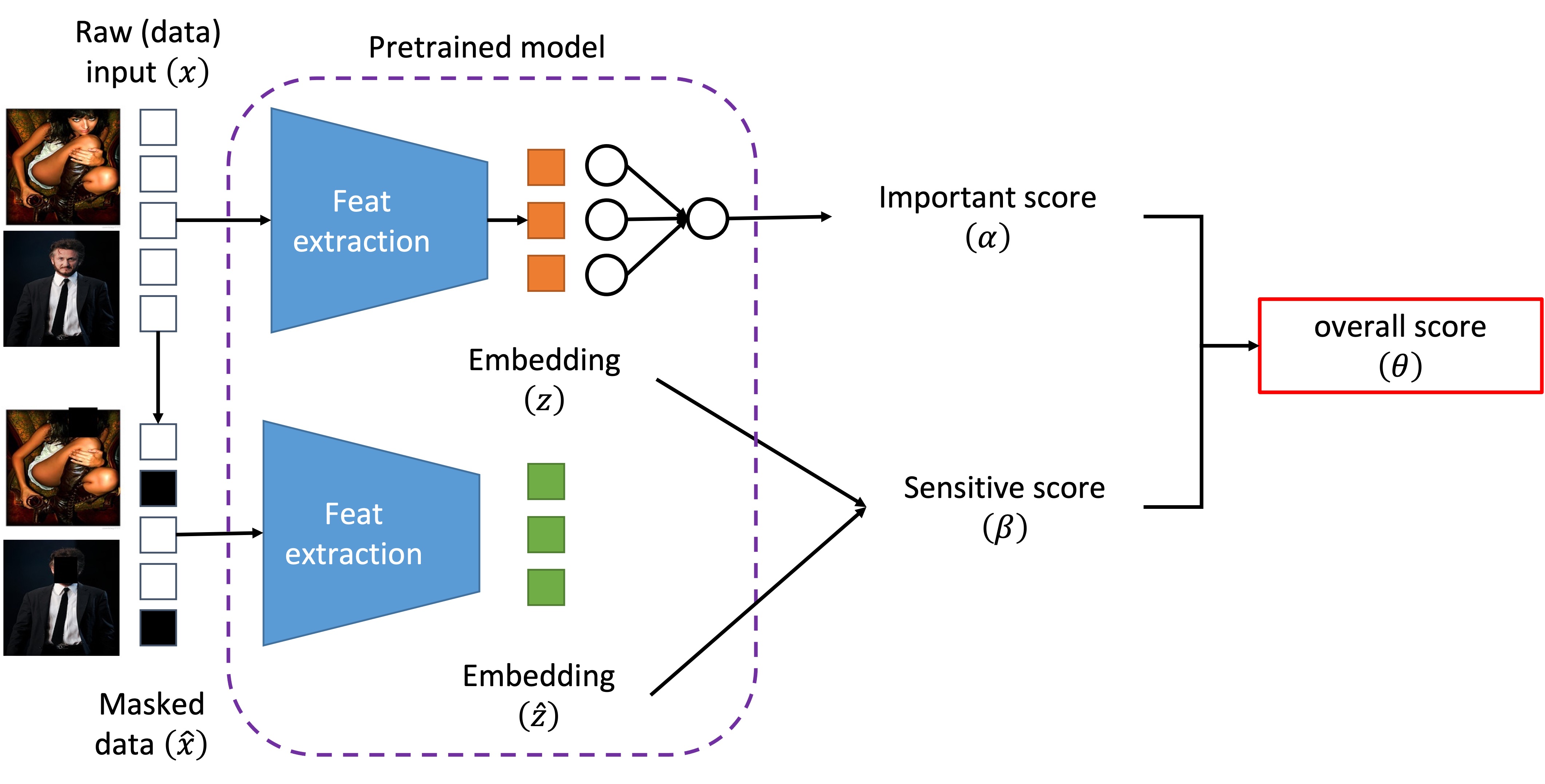} \vspace{-10pt}
    \caption{An overview of feature's importance and sensitivity indication process.}
    \vspace{-15pt}
    \label{fig:impsensextract}
\end{figure}

\begin{figure}[h]
    \centering\includegraphics[width=0.3\columnwidth]{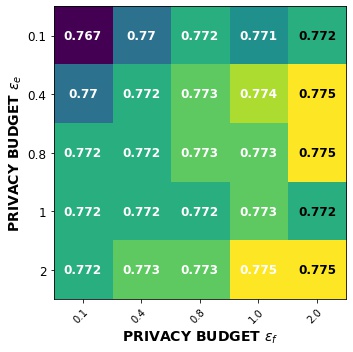} \vspace{-10pt}
    \caption{Additional results of model performance of the combination of feature-level and edge-level protection of the FLICKR-MIR dataset.} \vspace{-5pt}
    \label{fig:resBRandERand-MIR}
\end{figure}

\begin{figure}[h]
% % \vskip 0.2in
\centering     %%% not \center
    \subfigure[PPI with $\rho = 0.05$ (F1-SCORE)]{\includegraphics[width=0.3\columnwidth]{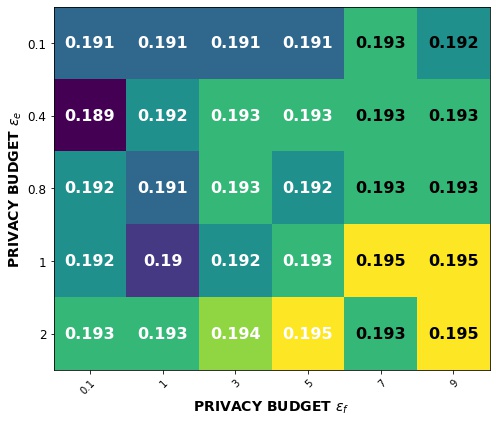}}
    \hfill
    \subfigure[PPI with $\rho=0.1$ (F1-SCORE)]{\includegraphics[width=0.3\columnwidth]{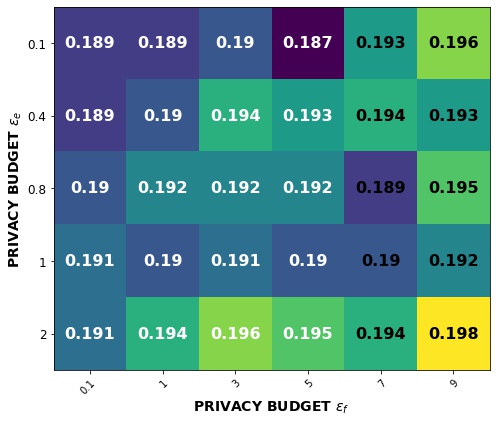}}
    \hfill
    \subfigure[PPI with $\rho=0.2$ (F1-SCORE)]{\includegraphics[width=0.3\columnwidth]{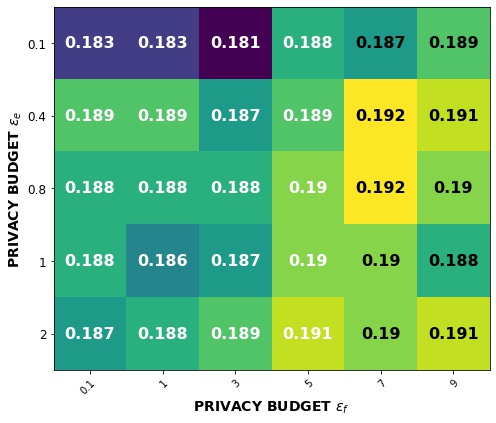}}
    \caption{Additional results of model performance of the combination of feature-level and edge-level protection for the PPI dataset.}
    \label{fig:sup_ppifeatedge}
\end{figure}

\end{document}